\documentclass{article} 

\usepackage{iclr2023_conference,times}

\usepackage{hyperref}
\usepackage{url}

\iclrfinalcopy 
\usepackage{amsmath} 
\usepackage{amssymb} 
\usepackage{multirow}
\usepackage{makecell} 
\usepackage[pdftex]{graphicx} 
\usepackage{subcaption} 
\usepackage{bm} 
\usepackage{stmaryrd} 
\usepackage{dsfont} 
\usepackage{pifont} 
\usepackage{ifthen} 
\usepackage{booktabs} 
\usepackage{soul}
\usepackage{wrapfig} 
\usepackage{colortbl} 

\definecolor{mediumtealblue}{rgb}{0.0, 0.33, 0.71}
\definecolor{darkpastelgreen}{rgb}{0.01, 0.75, 0.24}
\definecolor{azure}{rgb}{0.0, 0.5, 1.0}

\newcommand{\D}{\mathsf{D}}
\newcommand{\G}{\mathsf{G}}
\newcommand{\F}{\mathsf{F}}
\newcommand{\A}{\mathsf{A}}
\newcommand{\C}{\mathsf{C}}
\newcommand{\E}{\mathsf{E}}
\newcommand{\T}{\mathsf{T}}

\newcommand{\Ss}{\mathsf{S}}
\newcommand{\ours}{\textcolor{azure}{(ours)}}

\newcommand{\R}{\mathbb{R}}

\newcommand{\M}{\mathsf{M}}

\newcommand{\campos}{\bm{\varphi}_\text{pos}}
\newcommand{\camfov}{\varphi_\text{fov}}
\newcommand{\camlookat}{\bm{\varphi}_\text{lookat}}
\newcommand{\camresidual}{\Delta \bm{\varphi}}
\newcommand{\camall}{\bm{\varphi}}
\newcommand{\camallprior}{\bm{\varphi}'}
\newcommand{\camprior}{\varphi_i'}
\newcommand{\camnew}{\varphi_i}

\newcommand{\depth}{\bm d}
\newcommand{\depthnorm}{\bar{\bm d}}
\newcommand{\depthadapt}{\bm d_a}
\newcommand{\depthone}{\bm d_a^{(1)}}
\newcommand{\depthtwo}{\bm d_a^{(2)}}
\newcommand{\depththree}{\bm d_a^{(3)}}
\newcommand{\depthreal}{\bm d_r}

\newcommand{\dogs}{SDIP Dogs}
\newcommand{\elephants}{SDIP Elephants}
\newcommand{\horses}{LSUN Horses}
\newcommand{\fidtwok}{FID$_\text{2k}$}

\newcommand{\dogsf}{\dogs$_\text{40k}$}
\newcommand{\horsesf}{\horses$_\text{40k}$}

\newcommand{\depthprob}{P(\depthnorm)}

\newcommand{\Ladv}{\mathcal{L}_\text{adv}}
\newcommand{\Lphi}{\mathcal{L}_{\varphi_i}}
\newcommand{\LG}{\mathcal{L}_\G}
\newcommand{\LD}{\mathcal{L}_\D}
\newcommand{\Ldist}{\mathcal{L}_\text{dist}}
\newcommand{\Ltwo}{\mathcal{L}_2}

\newcommand{\Rpenalty}{\mathcal{R}_1}
\newcommand{\apref}[1]{\ref{#1}}

\newcommand{\modelfullname}{3D generator with Generic Priors}
\newcommand{\modelname}{3DGP} 
\newcommand{\gradpen}{Camera Gradient Penalty}

\newcommand{\figref}[1]{Fig.~\ref{#1}}
\newcommand{\secref}[1]{\S\ref{#1}}
\newcommand{\tabref}[1]{Tab.~\ref{#1}}

\newcommand{\projecturl}{https://snap-research.github.io/3dgp}
\newcommand{\projecthref}{\href{\projecturl}{\projecturl}}
\newcommand{\leresrepourl}{https://github.com/compphoto/BoostingMonocularDepth}

\newcommand{\metricfullname}{Non-Flatness Score}
\newcommand{\metricname}{NFS}

\newcommand{\expect}[2][]{
\ifthenelse{\equal{#1}{}}{
\mathbb{E}\left[#2\right]
}{
\underset{#1}{\mathbb{E}}\left[#2\right]
}}

\title{3D generation on ImageNet}

\author{Ivan Skorokhodov\thanks{Work done during internship at Snap Inc.} \\
KAUST
\And
Aliaksandr Siarohin \\ Snap Inc.
\And
Yinghao Xu$^*$ \\ CUHK
\And
Jian Ren \\ Snap Inc.
\And
Hsin-Ying Lee \\ Snap Inc.
\And
Peter Wonka \\ KAUST
\And
Sergey Tulyakov \\ Snap Inc.
}

\begin{document}

\maketitle

\begin{abstract}

All existing 3D-from-2D generators are designed for well-curated single-category datasets, where all the objects have (approximately) the same scale, 3D location and orientation, and the camera always points to the center of the scene.
This makes them inapplicable to diverse, in-the-wild datasets of non-alignable scenes rendered from arbitrary camera poses.
In this work, we develop \textit{\modelfullname\ (\modelname)}: a 3D synthesis framework with more general assumptions about the training data, and show that it scales to very challenging datasets, like ImageNet.
Our model is based on three new ideas.
First, we incorporate an \textit{inaccurate} off-the-shelf depth estimator into 3D GAN training via a special depth adaptation module to handle the imprecision.
Then, we create a flexible camera model and a regularization strategy for it to learn its distribution parameters during training.
Finally, we extend the recent ideas of transferring knowledge from pretrained classifiers into GANs for patch-wise trained models by employing a simple distillation-based technique on top of the discriminator.
It achieves more stable training than the existing methods and speeds up the convergence by at least 40\%.
We explore our model on four datasets: SDIP Dogs $256^2$, SDIP Elephants $256^2$, LSUN Horses $256^2$, and ImageNet $256^2$ and demonstrate that 3DGP outperforms the recent state-of-the-art in terms of both texture and geometry quality.

\begin{center}
Code and visualizations: \projecthref
\end{center}

\end{abstract}

\section{Introduction}\label{sec:introduction}

\begin{figure}[!h]
\centering
\includegraphics[width=\textwidth]{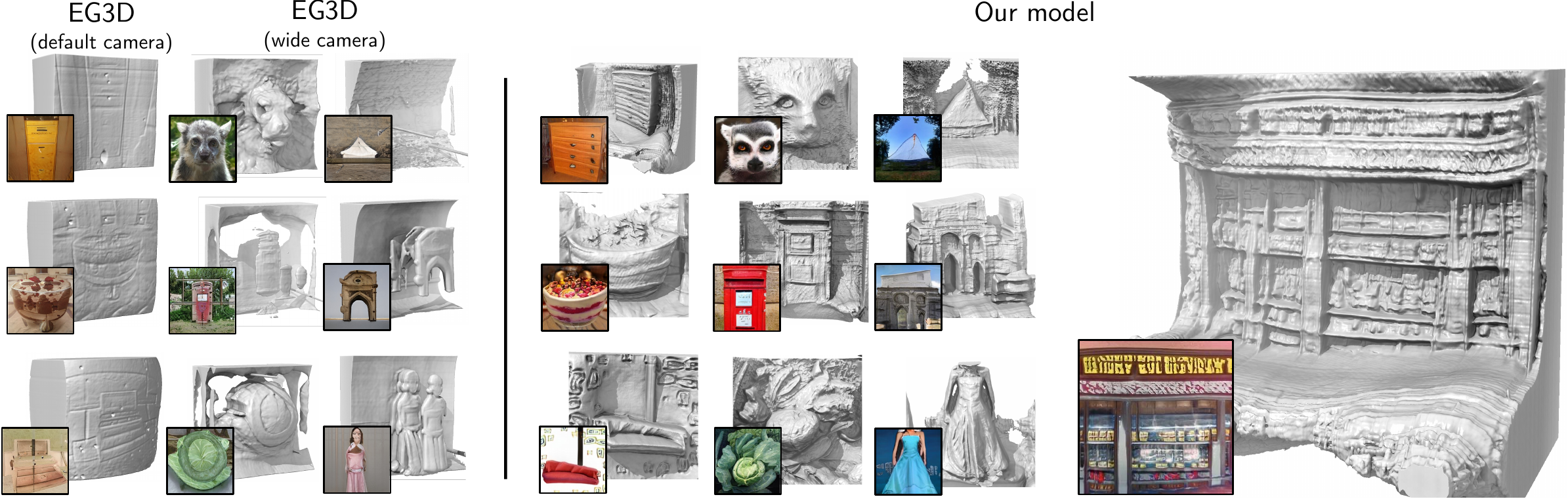}
    \caption{Selected samples from EG3D~\citep{EG3D} and our generator trained on ImageNet $256^2$~\citep{ImageNet}. EG3D models the geometry in low resolution and renders either flat shapes (when trained with the default camera distribution) or repetitive ``layered'' ones (when trained with a wide camera distribution). In contrast, our model synthesizes the radiance field in the full dataset resolution and learns high-fidelity details during training. Zoom-in for a better view.}
\label{fig:teaser}
\end{figure}

We witness remarkable progress in the domain of 3D-aware image synthesis.
The community is developing new methods to improve the image quality, 3D consistency and efficiency of the generators (e.g., ~\cite{EG3D, GRAM, EpiGRAF, GMPI, VoxGRAF}).
However, all the existing frameworks are designed for well-curated and aligned datasets consisting of objects of the same category, scale and global scene structure, like human or cat faces~\citep{piGAN}.
Such curation requires domain-specific 3D knowledge about the object category at hand, since one needs to infer the underlying 3D keypoints to properly crop, rotate and scale the images~\citep{GRAM,EG3D}.
This makes it infeasible to perform a similar alignment procedure for large-scale multi-category datasets that are inherently ``non-alignable'': there does not exist a single canonical position which all the objects could be transformed into (e.g., it is impossible to align a landscape panorama with a spoon).

\begin{figure}
\centering
\includegraphics[width=\textwidth]{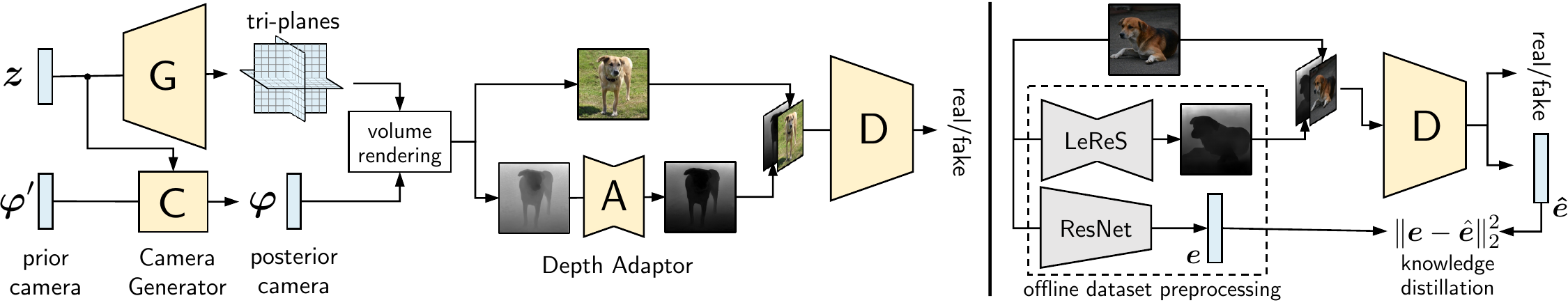}
\caption{\textbf{Model overview}. Left: our tri-plane-based generator. To synthesize an image, we first sample camera parameters from a prior distribution and pass them to the camera generator. This gives the posterior camera parameters, used to render an image and its depth map. The depth adaptor mitigates the distribution gap between the rendered and the predicted depth. Right: our discriminator receives a 4-channel color-depth pair as an input. A fake sample consists of the RGB image and its (adapted) depth map. A real sample consists of a real image and its estimated depth.
Our two-headed discriminator predicts adversarial scores and image features for knowledge distillation.
}
\label{fig:architecture-overview}
\vspace{-0.5cm}
\end{figure}

To extend 3D synthesis to in-the-wild datasets, one needs a framework which relies on more universal 3D priors.
In this work, we make a step towards this direction and develop a 3D generator with Generic Priors (\modelname): a 3D synthesis model which is guided only by (imperfect) depth predictions from an off-the-shelf monocular depth estimator.
Surprisingly, such 3D cues are enough to learn reasonable scenes from loosely curated, non-aligned datasets, such as ImageNet~\citep{ImageNet}.

Training a 3D generator on in-the-wild datasets comes with three main challenges: 1) extrinsic camera parameters of real images are unknown and impossible to infer; 2) objects appear in different shapes, positions, rotations and scales, complicating the learning of the underlying geometry; and 3) the dataset typically contains a lot of variation in terms of texture and structure, and is difficult to fit even for 2D generators.
As shown in Fig~\ref{fig:teaser} (left), state-of-the-art 3D-aware generators, such as EG3D~\citep{EG3D}, struggle to learn the proper geometry in such a challenging scenario.
In this work, we develop three novel techniques to address those problems.



\emph{Learnable ``Ball-in-Sphere" camera distribution}.
Most existing methods utilize a restricted camera model (e.g., ~\citep{GRAF, Giraffe, piGAN}): the camera is positioned on a sphere with a constant radius, always points to the world center and has fixed intrinsics.
But diverse, non-aligned datasets violate these assumptions: e.g., dogs datasets have images of both close-up photos of a snout and photos of full-body dogs, which implies the variability in the focal length and look-at positions.
Thus, we introduce a novel camera model with 6 degrees of freedom to address this variability.
We optimize its distribution parameters during training and develop an efficient gradient penalty for it to prevent its collapse to a delta distribution.

\emph{Adversarial depth supervision (ADS).}
A generic image dataset features a wide diversity of objects with different shapes and poses.
That is why learning a meaningful 3D geometry together with the camera distribution is an ill-posed problem, as the incorrect scale can be well compensated by an incorrect camera model~\citep{hartley2003multiple}, or flat geometry~\citep{GMPI,EG3D}.
To instill the 3D bias, we provide the scene geometry information to the discriminator by concatenating the depth map of a scene as the 4-th channel of its RGB input.
For real images, we use their (\emph{imperfect}) estimates from a generic off-the-shelf monocular depth predictor~\citep{LeReS}.
For fake images, we render the depth from the synthesized radiance field, and process it with a shallow depth adaptor module, bridging the distribution gap between the estimated and rendered depth maps.
This ultimately guides the generator to learn the proper 3D geometry.



\emph{Knowledge distillation into Discriminator.}
Prior works found it beneficial to transfer the knowledge from off-the-shelf 2D image encoders into a synthesis model~\citep{StyleGAN-XL}.
They typically utilize pre-trained image classifiers as the discriminator backbone with additional regularization strategies on top~\citep{ProjectedGANs, EnsemblingOffTheShelfModels}.
Such techniques, however, are only applicable when the discriminator has an input distribution similar to what the encoder was trained on.
This does not suit the setup of patch-wise training~\citep{GRAF} or allows to provide depth maps via the 4-th channel to the discriminator.
That is why we develop a more general and efficient knowledge transfer strategy based on knowledge distillation.
It consists in forcing the discriminator to predict features of a pre-trained ResNet50~\citep{ResNet} model, effectively transferring the knowledge into our model.
This technique has just 1\% of computational overhead compared to standard training, but allows to improve FID for both 2D and 3D generators by at least 40\%. 

First, we explore our ideas on \emph{non-aligned} single-category image datasets: \dogs~ $256^2$~\citep{SDIP}, \elephants~$256^2$~\citep{SDIP}, and \horses~$256^2$~\citep{LSUN}.
On these datasets, our generator achieves better image appearance (measured by FID~\cite{FID}) and geometry quality than the modern state-of-the-art 3D-aware generators.
Then, we train the model on all the {$1,000$} classes of ImageNet~\citep{ImageNet}, showing that multi-categorical 3D synthesis is possible for non-alignable data (see \figref{fig:teaser}).

\section{Related work}

\label{sec:related-work}

\textbf{3D-aware image synthesis}. 
\cite{NeRF} introduced Neural Radiance Fields (NeRF): a neural network-based representation of 3D volumes which is learnable from RGB supervision only.
It ignited many 3D-aware image/video generators~\citep{GRAF, Giraffe,piGAN,GIRAFFE-HD,CIPS-3D,CLIP-NeRF,StyleNeRF,StyleSDF,EG3D,EpiGRAF,MVCGAN,VolumeGAN,3D_video_gen}, all of them being GAN-based~\citep{GANs}.
Many of them explore the techniques to reduce the cost of 3D-aware generation for high-resolution data, like patch-wise training (e.g., \cite{GRAF, GNeRF, EpiGRAF}), MPI-based rendering (e.g., \cite{GMPI}) or training a separate 2D upsampler (e.g.,~\cite{StyleNeRF}).

\textbf{Learning the camera poses}.
NeRFs require known camera poses, obtained from multi-view stereo~\citep{schoenberger2016mvs} or structure from motion~\citep{SFM}.
Alternatively, a group of works has been introduced to either automatically estimate the camera poses~\citep{NeRF--} or finetune them during training~\citep{BaRF, NeROIC}.
The problem we are tackling in this work is fundamentally different as it requires learning not the camera poses from multi-view observations, but a \emph{distribution} of poses, while having access to sparse, single-view data of diverse object categories.
In this respect, the work closest to ours is CAMPARI~\citep{CAMPARI} as it also learns a camera distribution.

\textbf{GANs with external knowledge.}
Several works observed improved convergence and fidelity of GANs when using existing, generic image-based models~\citep{EnsemblingOffTheShelfModels, ProjectedGANs, StyleGAN-XL, FreezeD}, the most notable being StyleGAN-XL~\citep{StyleGAN-XL}, which uses a pre-trained EfficientNet~\citep{EfficientNet} followed by a couple of discriminator layers.
A similar technique is not suitable in our case as pre-training a generic RGB-D network on a large-scale RGB-D dataset is problematic due to the lack of data.
Another notable example is FreezeD~\cite{FreezeD}, which proposes to distill discriminator features for GAN finetuning.
Our work, on the other hand, relies on an existing model for image classification.






\textbf{Off-the-shelf depth guidance}.
GSN~\citep{GSN} also concatenates depth maps as the 4-th channel of the discriminator's input, but they utilize ground truth depths, which are not available for large-scale datasets.
DepthGAN~\citep{DepthGAN} uses predictions from a depth estimator to guide the training of a 2D GAN.
Exploiting monocular depth estimators for improving neural rendering was also explored in concurrent work~\citep{MonoSDF}, however, their goal is just geometry reconstruction. 
The core characteristic of our approach is taking the depth estimator imprecision into account by training a depth adaptor module to mitigate it (see \secref{sec:method:depth-adaptor}).

\section{Method}\label{sec:method}

We build our generator on top of EpiGRAF~\citep{EpiGRAF} since its fast to train, achieves reasonable image quality and does not need a 2D upsampler, relying on patch-wise training instead~\citep{GRAF}.
Given a random latent code $\bm z$, our generator $\G$ produces a tri-plane representation for the scene.
Then, a shallow 2-layer MLP predicts $\text{RGB}$ color and density $\sigma$ values from an interpolated feature vector at a 3D coordinate.
Then, images and depths are volumetrically rendered~\citep{NeRF} at any given camera position.
Differently to prior works~\citep{piGAN, Giraffe} that utilize fixed camera distribution, we sample camera from a trainable camera generator $\C$ (see \secref{sec:method:camera}).
We render depth and process it via the depth adaptor (see \secref{sec:method:depth-adaptor}), bridging the domains of rendered and estimated depth maps~\citep{LeReS}.
Our discriminator $\D$ follows the architecture of StyleGAN2~\citep{StyleGAN2-ADA}, additionally taking either adapted or estimated depth as the 4-th channel. 
To further improve the image fidelity, we propose a simple knowledge distillation technique, that enriches $\D$ with external knowledge obtained from ResNet~\citep{ResNet} (see \secref{sec:method:distillation}).
The overall model architecture is shown in \figref{fig:architecture-overview}.

\subsection{Learnable ``Ball-in-Sphere'' camera distribution}\label{sec:method:camera}

\begin{wrapfigure}{r}{7cm}
\centering
\begin{subfigure}[b]{3.4cm}
    \centering
    \includegraphics[width=\linewidth]{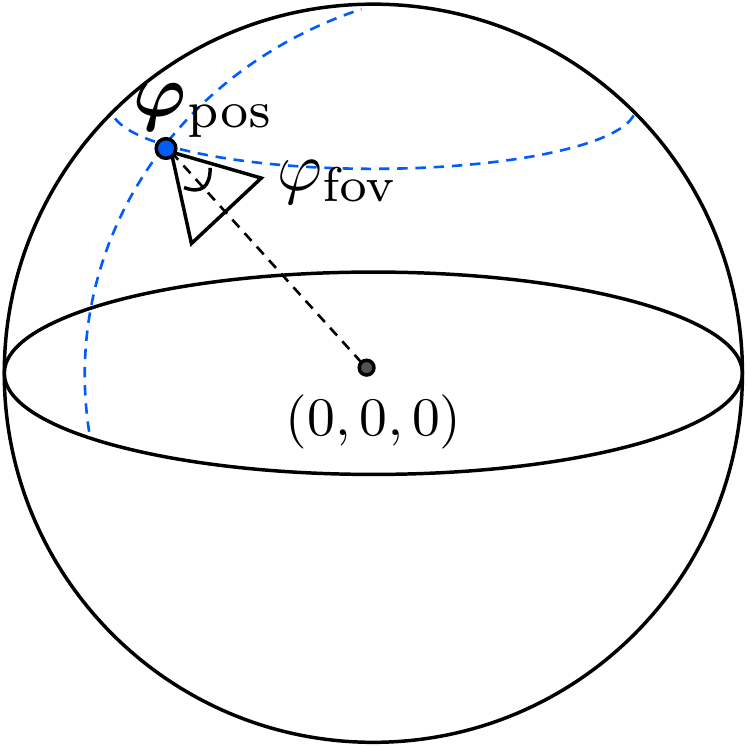}
    \hfill
    \caption{Standard camera model}
    \label{fig:standard-camera}
\end{subfigure}
\hfill
\begin{subfigure}[b]{3.4cm}
    \centering
    \includegraphics[width=\linewidth]{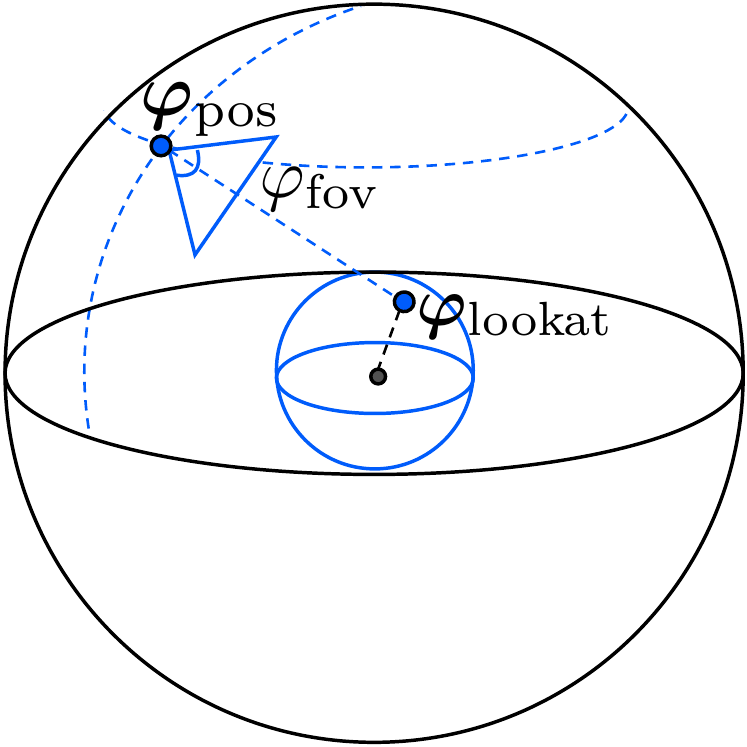}
    \hfill
    \caption{Our camera model}
    \label{fig:our-camera}
\end{subfigure}
\caption{\textbf{Camera model}.
(a) Conventional camera model is designed for aligned datasets and uses just 2 degrees of freedom.
(b) The proposed ``Ball-in-Sphere" parametrization has 4 additional degrees of freedom: field of view and the look at position. 
Variable parameters are shown in blue.
}
\vspace{-15pt}
\label{fig:camera-parametrization}
\end{wrapfigure}

\textbf{Limitations of Existing Camera Parameterization.}
The camera parameterization of existing 3D generators follows an overly simplified distribution --- its position is sampled on a fixed-radius sphere with fixed intrinsics, and the camera always points to $(0, 0, 0)$. This parametrization has only two degrees of freedom: pitch and yaw ($\campos$ in \figref{fig:camera-parametrization} (a)), implicitly assuming that all the objects could be centered, rotated and scaled with respect to some canonical alignment.
However, natural in-the-wild 3D scenes are inherently non-alignable: they could consist of multiple objects, objects might have drastically different shapes and articulation, or they could even be represented only as volumes (like smoke).
This makes the traditional camera conventions ill-posed for such data.

\textbf{Learnable ``Ball-in-Sphere'' Camera Distribution.}
We introduce a new camera parametrization which we call ``Ball-in-Sphere''.
Contrary to the standard one, it has four additional degrees of freedom: the field of view $\camfov$, and pitch, yaw, and radius of the inner sphere, specifying the look-at point within the outer sphere ($\camlookat$ in \figref{fig:camera-parametrization} (b)).
Combining with the standard parameters on the outer sphere, our camera parametrization has six degrees of freedom $\camall = \left[\campos \mathbin\Vert \camfov \mathbin\Vert \camlookat\right]$, where $\mathbin\Vert$ denotes concatenation.

Instead of manually defining camera distributions, we learn the camera distribution during training for each dataset.
In particular, we train the camera generator network $\C$ that takes camera parameters sampled from a sufficiently wide camera prior $\camallprior$ and produces new camera parameters $\camall$. For a class conditional dataset, such as ImageNet where scenes have significantly different geometry, we additionally condition this network on the class label $c$ and the latent code $\bm z$, i.e. $\camall = \C(\camallprior, \bm z, c)$. For a single category dataset we use $\camall = \C(\camallprior, \bm z)$.

\textbf{{\gradpen}.}
To the best of our knowledge, CAMPARI~\citep{CAMPARI} is the only work which also learns the camera distribution. It samples a set of camera parameters from a wide distribution and passes it to a shallow neural network, which produces a residual $\camresidual=\camall - \camallprior$ for these parameters.
However, we observed that such a regularization is too weak for the complex datasets we explore in this work, and leads to a distribution collapse (see \figref{fig:camera-reg-comparison}).
Note, that a similar observation was also made by~\cite{StyleNeRF}.

To prevent the camera generator from producing collapsed camera parameters, we seek a new regularization strategy. Ideally, it should prevent constant solutions, while at the same time, reducing the Lipschitz constant for $\C$, which is shown to be crucial for stable training of generators~\citep{JacobianForGenerator}. We can achieve both if we push the derivatives of the predicted camera parameters with respect to the prior camera parameters to either one or minus one, arriving at the following regularization term: 
\begin{equation}\label{eq:camera-gen-grad-penalty}
\Lphi = \left\vert \frac{\partial \camnew}{\partial \camprior} \right\vert + \left\vert \frac{\partial \camnew}{\partial \camprior} \right\vert ^ {-1},
\end{equation}
where $\camprior \in \camallprior$ is the camera sampled from the prior distribution and $\camnew \in \camall$ is produced by the camera generator.
We refer to this loss as {\gradpen}.
Its first part prevents rapid camera changes, facilitating stable optimization, while the second part avoids collapsed posteriors.

\begin{figure}
\centering
\includegraphics[width=\textwidth]{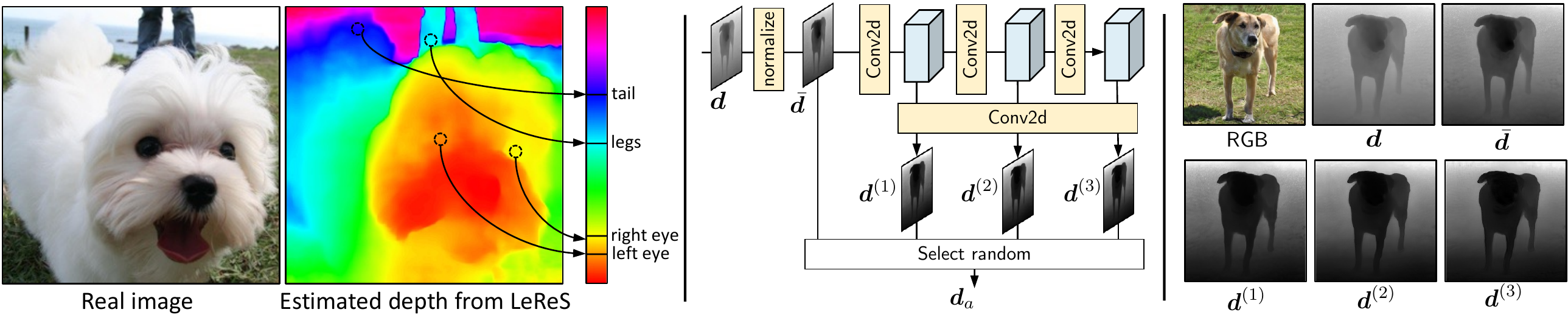}
\caption{\textbf{Depth adapter.} Left: An example of a real image with its depth estimated by LeReS~\citep{LeReS}.  Note that the estimated depth has several artifacts. For example, the human legs are closer than the tail, eyes are spaced unrealistically, {and} far-away grass is predicted to be close. Middle: depth adapter meant to bridge the domains of predicted and NeRF-rendered depth. Right: a generated image with its adapted depth maps obtained from different layers of the adapter.}
\label{fig:depth-adaptor}
\end{figure}

\subsection{Adversarial depth supervision}\label{sec:method:depth-adaptor}

To instill a 3D bias into our model, we develop a strategy of using depth maps predicted by an off-the-shelf estimator
$\E$~\citep{LeReS}, for its advantages of being generic and readily applicable for many object categories.
The main idea is concatenating a depth map as the 4-th channel of the RGB as the input of the discriminator. The fake depth maps in this case is obtained with the help of neural rendering, while the real depth maps are estimated using monocular depth estimator $\E$.
However, naively utilizing the depth from $\E$ leads to training divergence.
This happens because $\E$ could only produce relative depth, not  metric depth. Moreover, monocular depth estimators are still not perfect, they produce noisy artifacts, ignore high-frequency details, and make prediction mistakes. Thus, we devise a mechanism that allows utilization of the imperfect depth maps. The central part of this mechanism is a learnable depth adaptor $\A$, that should transform and augment the depth map obtained with neural rendering to look like a depth map from $\E$.

More specifically, we first render raw depths $\depth$ from NeRF via volumetric rendering:
\begin{equation}
\depth = \int_{t_n}^{t_f} T(t) \sigma (r(t)) t dt,
\label{eq:depth_rendering}
\end{equation}
where $t_n, t_f \in \R$ are near/far planes, $T(t)$ is accumulated transmittance, and $r(t)$ is a ray. Raw depth is shifted and scaled from the range of $[t_n, t_f]$ into $[-1, 1]$ to obtain normalized depth $\depthnorm$:
\begin{equation}
\depthnorm = 2 \cdot \frac{\depth - (t_n + t_f + b) / 2}{t_f - t_n - b},
\end{equation}
where $b \in [0, (t_n + t_f) / 2]$ is an additional learnable shift needed to account for the empty space in the front of the camera.
Real depths are normalized into the $[-1, 1]$ range directly.

\begin{figure}
\centering
\includegraphics[width=\textwidth]{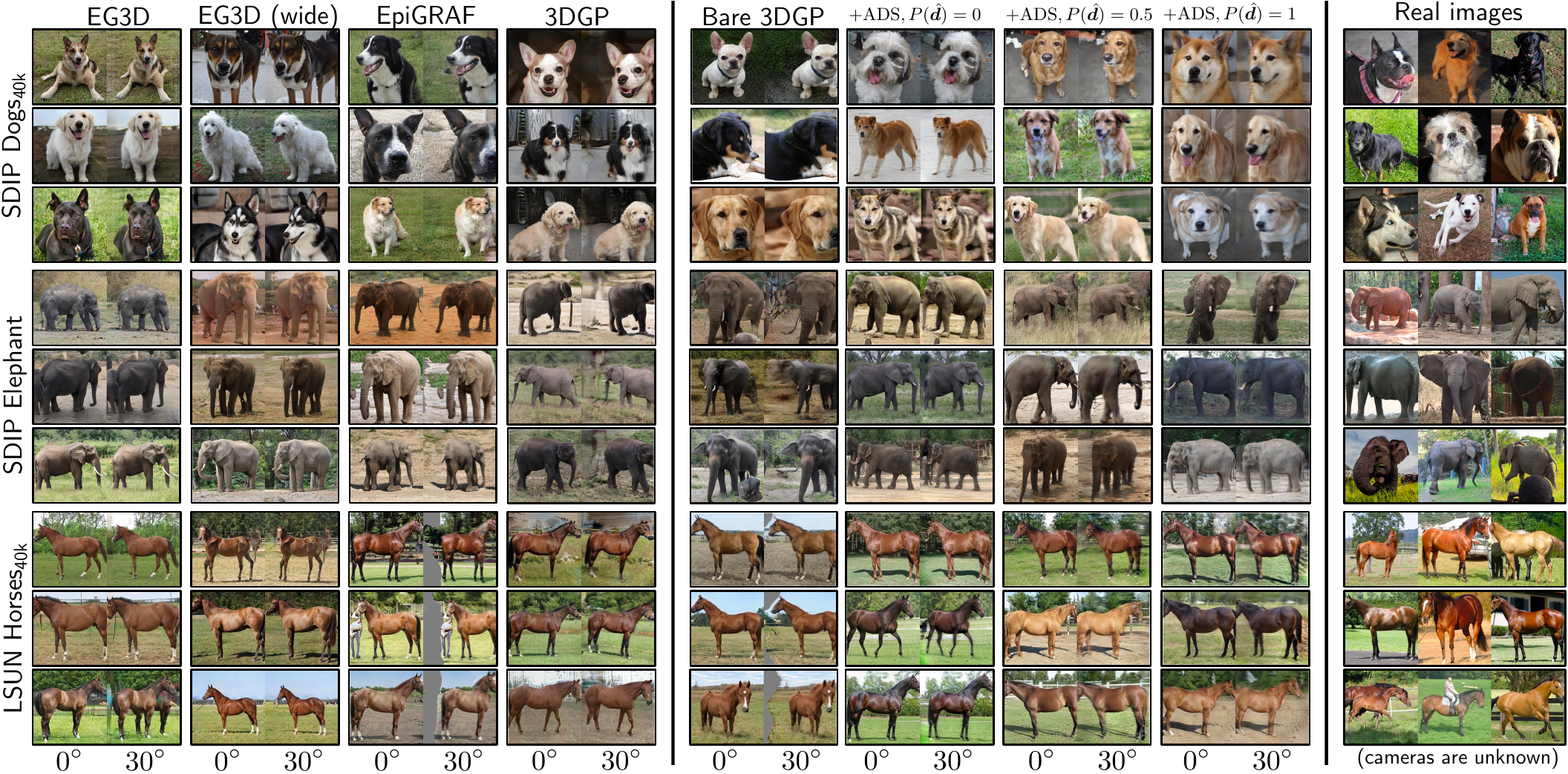}
\caption{
\textbf{Qualitative multi-view comparisons.}
Left: samples from the models trained on single-category datasets with articulated geometry.
Two views are shown for each sample. Middle: ablations following Tab.~\ref{tab:2d-experiments}, where we change the probability of using the normalized rendered depth $\depthprob$. 
EG3D, EpiGRAF, and $\depthprob=0$ do not render realistic side views, due to the underlying flat geometry.
Our full model instead generates realistic high-quality views on all the datasets.
Right: randomly sampled real images.
Zoom-in for greater detail.}
\label{fig:other-datasets}
\vspace{-0.5cm}
\end{figure}

\textbf{Learnable Depth Adaptor.} 
While $\depthnorm$ has the same range as $\depthreal$, it is not suitable for adversarial supervision directly due it the imprecision of $\E$: $\G$ would be trained to simulate all its prediction artifacts.
To overcome this issue, we introduce a \textit{depth adaptor} $\A$ to adapt the depth map $\depthadapt = \A(\depthnorm) \in \R^{h \times w}$, where $h \times w$ is a number of sampled pixels.
This depth (fake $\depthadapt$ or real $\depthreal$) is concatenated with the RGB input and passed to $\D$.

The depth adaptor $\A$ models artifacts produced by $\E$, so that the discriminator should focus only on the relevant high level geometry. However, a too powerful $\A$ would be able to fake the depth completely, and $\G$ will not learn the geometry.
This is why we structure $\A$ as just a 3-layer convolutional network (see \figref{fig:depth-adaptor}).
Each layer produces a separated depth map with different levels of adaptation: $\depthone, \depthtwo \text{and}~\depththree$.
The final adapted depth $\depthadapt$ is randomly selected from the set of  $\{\depthnorm, \depthone, \depthtwo, \depththree\}$.
Such design can effectively learn good geometry while alleviating overfitting.
For example, when $\D$ receives the original depth map $\depthnorm$ as input, it provides to $\G$ a strong signal for learning the geometry.
And passing an adapted depth map $\bm{d}^{(i)}_a$ to $\D$ allows $\G$ to simulate the imprecision artifacts of the depth estimator without degrading its original depth map $\depthnorm$.



\subsection{Knowledge distillation for Discriminator}\label{sec:method:distillation}

Knowledge from pretrained classification networks was shown to improve training stability and generation quality in 2D GANs~\citep{ProjectedGANs, EnsemblingOffTheShelfModels, StyleGAN-XL, InstanceConditionedGAN}. A popular solution proposed by~\cite{ProjectedGANs, StyleGAN-XL} is to use an off-the-shelf model as a discriminator while freezing most of its weights. Unfortunately, this technique is not applicable in our scenario since we modify the architecture of the discriminator by adding an additional depth input (see \secref{sec:method:depth-adaptor}) and condition on the parameters of the patch similarly to EpiGRAF~\citep{EpiGRAF}. Thus, we devise an alternative technique that can work with arbitrary architectures of the discriminator. 
Specifically, for each real sample{,} {we obtain two feature representations: $e$ from the pretrained ResNet~\citep{ResNet} network and $\hat{e}$ extracted from the final representation of our discriminator $\D$.}
Our loss simply pushes $\hat{e}$ to $e$ as follows:
\begin{equation}
\Ldist = \left\| \bm e - \hat{\bm e} \right\| ^2_2.
\end{equation}
$\Ldist$ can effectively distill knowledge from the pretrained ResNet~\citep{ResNet} into our $\D$.

\subsection{Training}\label{sec:method:training}

The overall loss for generator $\G$ consists of two parts: adversarial loss and {\gradpen}: 
\begin{equation}
\LG = \Ladv + \sum_{\camnew \in \camall} \lambda_{\camnew} \Lphi,
\end{equation}
where $\Ladv$ is {the} non-saturating loss~\citep{GANs}. We observe that a diverse distribution for camera origin is  most important for leaning meaningful geometry, but it is also most prone to degrade to a constant solution. 
Therefore, we set $\lambda_{\camnew}=0.3$ for $\campos$, while set $\lambda_{\camnew}=0.03$ for $\camfov$ and $\lambda_{\camnew}=0.003$ for $\camlookat$.
The loss for discriminator $\D$, on the other hand, consists of three parts: adversarial loss, knowledge distillation, and $\Rpenalty$ gradient penalty~\citep{R1_reg}:
\begin{equation}
\LD = \Ladv + \lambda_\text{dist} \Ldist + \lambda_r \Rpenalty.
\end{equation}
We use the same optimizer and hyper-parameters as EpiGRAF.
We observe that for depth adaptor{,} sampling adapted depth maps with equal probability is not always beneficial, and found that using $\depthprob = 0.5$ leads to better geometry.
For additional details, see Appx~\ref{ap:implementation-details}.

\section{Experimental Results}\label{sec:experiments}

\textbf{Datasets.}
In our experiments, we use 4 \emph{non-aligned} datasets: \dogs~\citep{SDIP}, \elephants~\citep{SDIP}, \horses~\citep{LSUN}, and ImageNet~\citep{ImageNet}.
The first three are single-category datasets that contain objects with complex articulated geometry, making them challenging for standard 3D generators.
We found it useful to remove outlier images from \dogs\ and \horses via instance selection~\citep{InstanceSelectionForGANs}, reducing their size to $40$K samples each.
We refer to the filtered versions of these datasets as \dogsf, and \horsesf, respectively.
We then validate our method on ImageNet, a real-world, multi-category dataset containing $1,000$ diverse object classes, with more than $1,000$ images per category.
All the 3D generators (including the baselines) use the same filtering strategy for ImageNet, where 2/3 of its images are filtered out.
Note that all the metrics are \emph{always} measured on the full ImageNet. 

\begin{wrapfigure}{r}{0.4\textwidth}
\begin{center}
\includegraphics[width=0.4\textwidth]{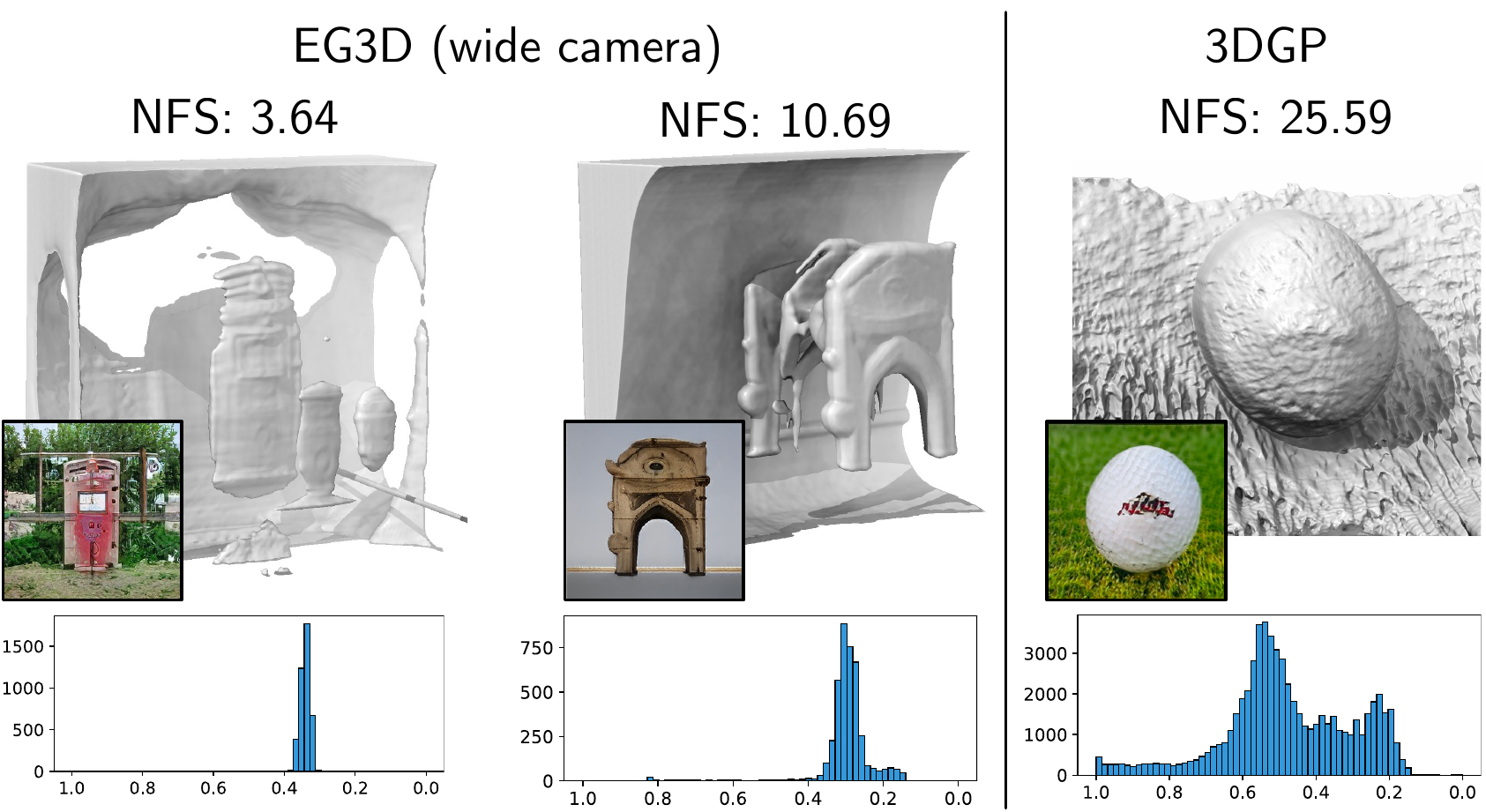}
\end{center}
\caption{{\metricname} for repetitive and diverse geometry with depth histograms.}
\vspace{-10pt}
\label{fig:nfs}
\end{wrapfigure}

\textbf{Evaluation.}
We rely on FID~\citep{FID} to measure the image quality and FID$_\text{2k}$, which is computed on 2,048 images instead of 50k (as for FID) for efficiency.
For ImageNet, we additionally compute Inception Score (IS)~\citep{InceptionScore}.
Note that while we train on the filtered ImageNet, we always compute the metrics on the full one.
There is no established protocol to evaluate the geometry quality of 3D generators in general, but the state-of-the-art ones are tri-plane~\citep{EG3D, IDE-3D} or MPI-based~\cite{GMPI}, and we observed that it is possible to quantify their most popular geometry failure case: flatness of the shapes.
For this, we propose Non-Flatness Score (\metricname) which is computed as the average entropy of the normalized depth maps histograms.
We depict its intuition in Fig~\ref{fig:nfs} and provide the details in Appx~\ref{ap:metric-details}.

\subsection{3D generation for single category datasets}

We show quantitative results for single category datasets in Tab.~\ref{tab:2d-experiments}a, where we compare with EG3D~\citep{EG3D} and EpiGRAF~\citep{EpiGRAF}.
EG3D was mainly designed for FFHQ~\citep{StyleGAN} and uses the true camera poses inferred from real images as the camera distribution for the generator.
But in our scenario, we do not have any knowledge about the true camera distribution, that is why to stay as close as possible to the setup EG3D was designed for, we use normal distribution for rotation and elevation angles with the same standard deviations as in FFHQ, which are equal to $\sigma_\text{yaw}=0.3$ and $\sigma_\text{pitch}=0.155$, respectively.
Also, to learn better geometry, we additionally trained the baselines with a twice wider camera distribution: $\sigma_\text{yaw}=0.6$ and $\sigma_\text{pitch}=0.3$.
While it indeed helped to reduce flatness, it also worsened the image quality: up to 500\% as measured by \fidtwok.
Our model shows substantially better (at least $2\times$ than EG3D, slightly worse than StyleGAN2) \fidtwok\ and greater \metricname~on all the datasets.
Low \metricname~indicates flat or repetitive geometry impairing the ability of the model to generate realistic side views.
Indeed, as shown in \figref{fig:other-datasets} (left), both EG3D (with the default camera distribution) and EpiGRAF struggle to generate side views, while our method (\modelname) renders realistic side views on all three datasets.

\begin{table*}[t]
\caption{Comparisons on \dogsf,~\elephants,~and~\horsesf.
(a) includes EG3D (with the standard and the wider camera range), EpiGRAF, and 3DGP. 
For completeness, we also provide 2D baseline StyleGAN2 (with KD). (b) includes the ablations of the proposed contributions.  
We report the total training cost for prior works, our model, and our proposed contributions.}
\label{tab:2d-experiments}
\centering
\resizebox{\linewidth}{!}{
\begin{tabular}{lcccccccc}

\multicolumn{8}{c}{ (a) Comparison of EG3D, EpiGRAF and \modelname.} \\
\toprule
\multirow{2}{*}{Model} & \multicolumn{2}{c}{SDIP Dogs$_\text{40k}$} & \multicolumn{2}{c}{SDIP Elephants$_\text{40k}$} & \multicolumn{2}{c}{LSUN Horses$_\text{40k}$} & Training cost \\
& FID$_\text{2k}$ $\downarrow$ & \metricname $\uparrow$ & FID$_\text{2k}$ $\downarrow$ & \metricname $\uparrow$ & FID$_\text{2k}$ $\downarrow$ & \metricname $\uparrow$ & (A100 days) \\
\midrule
EG3D & 16.2 & 11.91 & 4.78 & 2.59 & 3.12 & 13.34 & 3.7  \\
~+ wide camera & 21.1 & 24.44 & 5.76 & 17.88 & 19.44 & 25.34 & 3.7  \\
EpiGRAF & 25.6 & 3.53 & 8.24 & 12.9 & 6.45 & 9.73 & 2.3 \\ 
3DGP \ours & 8.74 & 34.35 & 5.79 & 32.8 & 4.86 & 30.4 & 2.6 \\ 
\midrule
StyleGAN2 (with KD) & 6.24 & N/A & 3.94 & N/A & 2.57 & N/A & 1.5 \\
\bottomrule
\\
\multicolumn{8}{c}{(b) Impact of Adversarial Depth Supervision (ADS).} \\
\toprule
Bare 3DGP (w/o ADS, w/o $\C$) & 8.59 & 1.42 & 7.46 & 9.52 & 3.29 & 8.04 & 2.3 \\
~+ ADS, $\depthprob = 0.0$ & 8.13 & 3.14 & 5.69 & 1.97 & 3.41 & 1.24 & 2.6 \\
~+ ADS, $\depthprob = 0.25$ & 9.57 & 33.21 & 6.26 & 33.5 & 4.33 & 32.68 & 2.6 \\
~+ ADS, $\depthprob = 0.5$ & 9.25 & 36.9 & 7.60 & 30.7 & 5.27 & 32.24 & 2.6 \\
~+ ADS, $\depthprob = 1.0$ & 12.2 & 27.2 & 12.1 & 26.0 & 8.24 & 27.7 & 2.5 \\
\bottomrule
\end{tabular}
}
\end{table*}

\textbf{Adversarial Depth Supervision.} 
We evaluate the proposed Adversarial Depth Supervision (ADS) and the depth adaptor $\A$.
The only hyperparameter it has is the probability of using the non-adapted depth $\depthprob$ (see \secref{sec:method:depth-adaptor}). We ablate this parameter in Tab.~\ref{tab:2d-experiments}b.
While FID scores are only slightly affected by varying $\depthprob$, we see substantial difference in the \metricfullname.
We first verify that \metricname~is the worst without ADS, indicating the lack of a 3D bias. When $\depthprob = 0$, the discriminator is never presented with the rendered depth $\depthnorm$, while the adaptor learns to fake the depth, leading to flat geometry.
When $\depthprob=1$, the adaptor is never used, allowing $\D$ to easily determine the generated depth from the estimated depth, as there is a large domain gap (see \figref{fig:depth-adaptor}), leading to reduced FID scores.
The best results are achieved with $\depthprob=0.5$, which can be visually verified from observing \figref{fig:other-datasets} (middle).
A ``bare'' \modelname, $\depthprob=0$ and $\depthprob=1$ are unable to render side views, while the model trained with $\depthprob=0.5$ features the overall best geometry and side views.

\begin{figure}
\centering
\includegraphics[width=\textwidth]{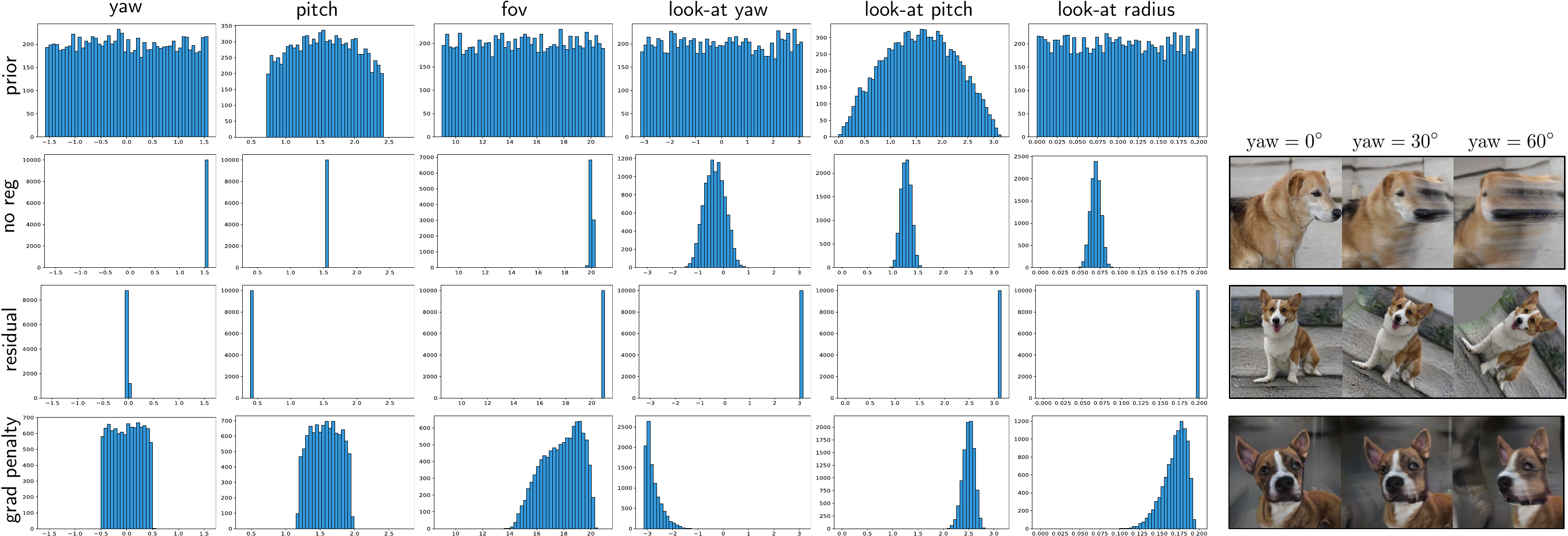}
\caption{\textbf{Comparisons of regularization strategies.} Left: selected generic and wide prior for each of the 6 DoFs of our camera model (top row). Without any regularization (\emph{no reg}) or with the residual-based model (\emph{residual}) as in~\cite{CAMPARI}, the camera generator collapses to highly concentrated distributions. In contrast, the proposed regularization (\emph{grad penalty}) leads to a wider posterior. Right: random samples for each strategy. For (\emph{no reg}) and (\emph{residual}), no meaningful geometry is learned; only our method (\emph{grad penalty}) leads to good geometry.}
\label{fig:camera-reg-comparison}
\vspace{-0.5cm}
\end{figure}

\textbf{Knowledge Distillation.}
Here we further provide insights for discriminator $\D$ using our knowledge distillation strategy (see \secref{sec:method:distillation}). We find that knowledge distillation provides an additional stability for adversarial training, along with the significant improvement in FID, which can be observed by comparing results of EpiGRAF in Tab.~\ref{tab:2d-experiments}a and Bare 3DGP Tab.~\ref{tab:2d-experiments}a. Additionally, we compare different knowledge distillation strategies in Appx~\ref{ap:knowledge-distillation}. However, as noted by ~\cite{FID_evaluation}, strategies which utilize additional classification networks, may provide a significant boost to FID, without corresponding improvement in visual quality.


\textbf{``Ball-in-Sphere" Camera Distribution.} 
In \figref{fig:camera-reg-comparison}, we analyze different strategies for learning the camera distribution: sampling the camera $\camall$ from the prior $\camallprior$ without learning, predicting the residual $\camall - \camallprior$, and using our proposed camera generator $\C$ with {\gradpen}.
For the first two cases, the learned distributions are nearly deterministic.
Not surprisingly, no meaningful side views can be generated as geometry becomes flat.
In contrast, our regularization provides sufficient regularization to $\C$, enabling it to converge to a sufficiently wide posterior, resulting into valid geometry and realistic side views.
After the submission, we found a simpler and more flexible camera regularization strategy through entropy maximization, which we discuss in Appx~\ref{ap:entropy-reg}.

\begin{table}
\caption{Comparison between different generators on ImageNet $256^2$ (note that \modelname\ relies on additional information in the form of depth supervision). Training cost is measured in A100 days.}
\label{tab:main-results}
\centering
\resizebox{1.0\linewidth}{!}{
\begin{tabular}{lccccc}
\toprule
Method & Synthesis type & FID $\downarrow$ & IS $\uparrow$ & \metricname $\uparrow$ & A100 days $\downarrow$ \\
\midrule
BigGAN~\citep{BigGAN} & 2D & 8.7 & 142.3 & N/A & 60 \\
StyleGAN-XL~\citep{StyleGAN-XL} & 2D & 2.30 & 265.1 & N/A & 163+ \\
ADM~\citep{ADM} & 2D & 4.59 & 186.7 & N/A & 458 \\
\midrule
EG3D~\citep{EG3D} & 3D-aware & 26.7 & 61.4 & 3.70 & 18.7 \\
~+ wide camera & 3D-aware & 25.6 & 57.3 & 9.83 & 18.7 \\
VolumeGAN~\citep{VolumeGAN} & 3D-aware & 77.68 & 19.56 & 22.69 & 15.17 \\
StyleNeRF~\citep{StyleNeRF} & 3D-aware & 56.54 & 21.80 & 16.04 & 20.55 \\
StyleGAN-XL + 3DPhoto~\citep{3DPhoto} & 3D-aware & 116.9 & 9.47 & N/A & 165+ \\
EpiGRAF~\citep{EpiGRAF} & 3D & 47.56 & 26.68 & 3.93 & 15.9 \\
~+ wide camera & 3D & 58.17 & 20.36 & 12.89 & 15.9 \\
\modelname~\ours & 3D & 19.71 & 124.8 & 18.49 & 28 \\
\bottomrule
\end{tabular}
\vspace{-0.6cm}
}
\end{table}

\subsection{3D synthesis on ImageNet}

ImageNet~\citep{ImageNet} is significantly more difficult than single-category datasets.
Following prior works, we trained \emph{all} the methods in the conditional generation setting~\cite{BigGAN}.
The quantitative results are presented in Tab.~\ref{tab:main-results}.
We also report the results of state-of-the-art 2D generators for reference: as expected, they show better FID and IS than 3D generators, since they do not need to learn geometry, are trained on larger compute, and had been studied for much longer.
Despite our best efforts to find a reasonable camera distribution, both EG3D and EpiGRAF produce flat or repetitive geometry, while {\modelname} produces geometry with rich details (see \figref{fig:teaser}).
StyleNeRF~\citep{StyleNeRF} and VolumeGAN~\cite{VolumeGAN}, trained for conditional ImageNet generation with the narrow camera distribution, substantially under-performed in terms of visual quality.
We hypothesize that the reason lies in their MLP/voxel-based NeRF backbones: they have a better 3D prior than tri-planes, but are considerably more expensive, which, in turn, requires sacrifices in terms of the generator's expressivity.

Training a 3D generator from scratch is not the only way to achieve 3D-aware synthesis: one can lift an existing 2D generator into 3D using the techniques from 3D photo ``inpainting''.
To test this idea, we generate 10k images with StyleGAN-XL (the current SotA on ImageNet) and then run 3DPhoto~\citep{3DPhoto} on them.\footnote{We computed FID on 10k images rather than 50k (as for other generators) due to the computational costs: parallelized inference of 3DPhoto~\citep{3DPhoto} on 10k images took 2 days on 8 V100s.}
This method first lifts a photo into 3D via a pre-trained depth estimator and then inpaints the holes with a separately trained GAN model to generate novel views.
This method works well for negligible camera variations, but starts diverging when the camera moves more than $10^\circ$.
In \tabref{tab:main-results}, we report its FID/IS scores with the narrow camera distribution: $\sigma_\text{yaw} \sim \mathcal{N}(0, 0.3)$ and $\sigma_\text{pitch} \sim \mathcal{N}(\pi/2, 0.15)$).
The details are in Appx~\apref{ap:3d-photo}.

\section{Conclusion}\label{sec:conclusion}

In this work, we present the first 3D synthesis framework for in-the-wild, multi-category datasets, such as ImageNet. 
We demonstrate how to utilize generic priors in the form of (\emph{imprecise}) monocular depth and latent feature representation to improve the visual quality and guide the geometry.
Moreover, we propose the new ``Ball-in-Sphere'' camera model with a novel regularization scheme that enables learning meaningful camera distribution.
On the other hand, our work still has several limitations, such as sticking background, lower visual quality compared to 2D generators, and no reliable quantitative measure of generated geometry.
Additional discussion is provided in Appx~\apref{ap:limitations}. 
This project consumed ${\approx}$12 NVidia A100 GPU years in total.

\section{Reproducibility Statement}\label{sec:reproducibility}

Most importantly, we will release 1) the source code and the checkpoints of our generator as a separate github repo; and 2) fully pre-processed datasets used in our work (with the corresponding extracted depth maps).
In \secref{sec:method} and \secref{sec:experiments} and Appx~\apref{ap:implementation-details}, and also in the figures throughout the text, we provided a complete list of architecture and optimization details needed to reproduce our results.
We are also open to provide any further details on our work in public and/or private correspondence.

\section{Ethics Statement}\label{sec:ethics}

There are two main ethical concerns around deep learning projects on synthesis: generation of fake content to mislead people (e.g. fake news or deepfakes\footnote{\href{https://en.wikipedia.org/wiki/Deepfake}{https://en.wikipedia.org/wiki/Deepfake}.}) and potential licensing, authorship and distribution abuse. This was recently discussed in the research community in the context of 2D image generation (by Stable Diffusion~\cite{LDM} and DALL-E~\cite{DALLE-2}) and code generation (by Github Copilot\footnote{\href{https://github.com/features/copilot}{https://github.com/features/copilot}.}).
While the current synthesis quality of our generator would still need to be improved to fool an attentive human observer, we still encourage the research community to discuss ideas and mechanisms for preventing abuse in the future.


\clearpage
\bibliography{iclr2023_conference}

\begin{thebibliography}{76}
\providecommand{\natexlab}[1]{#1}
\providecommand{\url}[1]{\texttt{#1}}
\expandafter\ifx\csname urlstyle\endcsname\relax
  \providecommand{\doi}[1]{doi: #1}\else
  \providecommand{\doi}{doi: \begingroup \urlstyle{rm}\Url}\fi

\bibitem[Bahmani et~al.(2022)Bahmani, Park, Paschalidou, Tang, Wetzstein,
  Guibas, Van~Gool, and Timofte]{3D_video_gen}
Sherwin Bahmani, Jeong~Joon Park, Despoina Paschalidou, Hao Tang, Gordon
  Wetzstein, Leonidas Guibas, Luc Van~Gool, and Radu Timofte.
\newblock 3d-aware video generation.
\newblock \emph{arXiv preprint arXiv:2206.14797}, 2022.

\bibitem[Brock et~al.(2018)Brock, Donahue, and Simonyan]{BigGAN}
Andrew Brock, Jeff Donahue, and Karen Simonyan.
\newblock Large scale gan training for high fidelity natural image synthesis.
\newblock \emph{arXiv preprint arXiv:1809.11096}, 2018.

\bibitem[Casanova et~al.(2021)Casanova, Careil, Verbeek, Drozdzal, and
  Romero-Soriano]{InstanceConditionedGAN}
Arantxa Casanova, Marlène Careil, Jakob Verbeek, Michal Drozdzal, and Adriana
  Romero-Soriano.
\newblock Instance-conditioned gan.
\newblock In \emph{Proceedings of the Neural Information Processing Systems
  Conference}, 2021.

\bibitem[Chan et~al.(2021)Chan, Monteiro, Kellnhofer, Wu, and Wetzstein]{piGAN}
Eric~R Chan, Marco Monteiro, Petr Kellnhofer, Jiajun Wu, and Gordon Wetzstein.
\newblock pi-gan: Periodic implicit generative adversarial networks for
  3d-aware image synthesis.
\newblock In \emph{Proceedings of the IEEE Conference on Computer Vision and
  Pattern Recognition}, 2021.

\bibitem[Chan et~al.(2022)Chan, Lin, Chan, Nagano, Pan, Mello, Gallo, Guibas,
  Tremblay, Khamis, Karras, and Wetzstein]{EG3D}
Eric~R. Chan, Connor~Z. Lin, Matthew~A. Chan, Koki Nagano, Boxiao Pan,
  Shalini~De Mello, Orazio Gallo, Leonidas Guibas, Jonathan Tremblay, Sameh
  Khamis, Tero Karras, and Gordon Wetzstein.
\newblock Efficient geometry-aware {3D} generative adversarial networks.
\newblock In \emph{Proceedings of the IEEE Conference on Computer Vision and
  Pattern Recognition}, 2022.

\bibitem[Chang et~al.(2015)Chang, Funkhouser, Guibas, Hanrahan, Huang, Li,
  Savarese, Savva, Song, Su, et~al.]{ShapeNet}
Angel~X Chang, Thomas Funkhouser, Leonidas Guibas, Pat Hanrahan, Qixing Huang,
  Zimo Li, Silvio Savarese, Manolis Savva, Shuran Song, Hao Su, et~al.
\newblock Shapenet: An information-rich 3d model repository.
\newblock \emph{arXiv preprint arXiv:1512.03012}, 2015.

\bibitem[{Dawson-Haggerty et al.}(2019)]{Trimesh}
{Dawson-Haggerty et al.}
\newblock trimesh.
\newblock \emph{https://github.com/mikedh/trimesh}, 2019.
\newblock URL \url{https://trimsh.org/}.

\bibitem[Deng et~al.(2009)Deng, Dong, Socher, Li, Li, and Fei-Fei]{ImageNet}
Jia Deng, Wei Dong, Richard Socher, Li-Jia Li, Kai Li, and Li~Fei-Fei.
\newblock Imagenet: A large-scale hierarchical image database.
\newblock In \emph{2009 IEEE conference on computer vision and pattern
  recognition}, 2009.

\bibitem[Deng et~al.(2022)Deng, Yang, Xiang, and Tong]{GRAM}
Yu~Deng, Jiaolong Yang, Jianfeng Xiang, and Xin Tong.
\newblock Gram: Generative radiance manifolds for 3d-aware image generation.
\newblock In \emph{IEEE Computer Vision and Pattern Recognition}, 2022.

\bibitem[Denninger et~al.(2019)Denninger, Sundermeyer, Winkelbauer, Zidan,
  Olefir, Elbadrawy, Lodhi, and Katam]{BlenderProc}
Maximilian Denninger, Martin Sundermeyer, Dominik Winkelbauer, Youssef Zidan,
  Dmitry Olefir, Mohamad Elbadrawy, Ahsan Lodhi, and Harinandan Katam.
\newblock Blenderproc.
\newblock \emph{arXiv preprint arXiv:1911.01911}, 2019.

\bibitem[DeVries et~al.(2020)DeVries, Drozdzal, and
  Taylor]{InstanceSelectionForGANs}
Terrance DeVries, Michal Drozdzal, and Graham~W Taylor.
\newblock Instance selection for gans.
\newblock \emph{Advances in Neural Information Processing Systems}, 2020.

\bibitem[DeVries et~al.(2021)DeVries, Bautista, Srivastava, Taylor, and
  Susskind]{GSN}
Terrance DeVries, Miguel~Angel Bautista, Nitish Srivastava, Graham~W. Taylor,
  and Joshua~M. Susskind.
\newblock Unconstrained scene generation with locally conditioned radiance
  fields.
\newblock \emph{arXiv}, 2021.

\bibitem[Dhariwal \& Nichol(2021)Dhariwal and Nichol]{ADM}
Prafulla Dhariwal and Alexander Nichol.
\newblock Diffusion models beat gans on image synthesis.
\newblock \emph{Advances in Neural Information Processing Systems},
  34:\penalty0 8780--8794, 2021.

\bibitem[Eftekhar et~al.(2021)Eftekhar, Sax, Malik, and Zamir]{Omnidata}
Ainaz Eftekhar, Alexander Sax, Jitendra Malik, and Amir Zamir.
\newblock Omnidata: A scalable pipeline for making multi-task mid-level vision
  datasets from 3d scans.
\newblock In \emph{ICCV}, 2021.

\bibitem[Flamary et~al.(2021)Flamary, Courty, Gramfort, Alaya, Boisbunon,
  Chambon, Chapel, Corenflos, Fatras, Fournier, Gautheron, Gayraud, Janati,
  Rakotomamonjy, Redko, Rolet, Schutz, Seguy, Sutherland, Tavenard, Tong, and
  Vayer]{POT}
R{\'e}mi Flamary, Nicolas Courty, Alexandre Gramfort, Mokhtar~Z. Alaya,
  Aur{\'e}lie Boisbunon, Stanislas Chambon, Laetitia Chapel, Adrien Corenflos,
  Kilian Fatras, Nemo Fournier, L{\'e}o Gautheron, Nathalie~T.H. Gayraud,
  Hicham Janati, Alain Rakotomamonjy, Ievgen Redko, Antoine Rolet, Antony
  Schutz, Vivien Seguy, Danica~J. Sutherland, Romain Tavenard, Alexander Tong,
  and Titouan Vayer.
\newblock Pot: Python optimal transport.
\newblock \emph{Journal of Machine Learning Research}, 22\penalty0
  (78):\penalty0 1--8, 2021.
\newblock URL \url{http://jmlr.org/papers/v22/20-451.html}.

\bibitem[Goodfellow et~al.(2014)Goodfellow, Pouget-Abadie, Mirza, Xu,
  Warde-Farley, Ozair, Courville, and Bengio]{GANs}
Ian Goodfellow, Jean Pouget-Abadie, Mehdi Mirza, Bing Xu, David Warde-Farley,
  Sherjil Ozair, Aaron Courville, and Yoshua Bengio.
\newblock Generative adversarial nets.
\newblock In \emph{Proceedings of the Neural Information Processing Systems
  Conference}, 2014.

\bibitem[Gu et~al.(2022)Gu, Liu, Wang, and Theobalt]{StyleNeRF}
Jiatao Gu, Lingjie Liu, Peng Wang, and Christian Theobalt.
\newblock Stylenerf: A style-based 3d aware generator for high-resolution image
  synthesis.
\newblock In \emph{International Conference on Machine Learning}, 2022.

\bibitem[Hartley \& Zisserman(2003)Hartley and Zisserman]{hartley2003multiple}
Richard Hartley and Andrew Zisserman.
\newblock \emph{Multiple view geometry in computer vision}.
\newblock Cambridge university press, 2003.

\bibitem[He et~al.(2016)He, Zhang, Ren, and Sun]{ResNet}
Kaiming He, Xiangyu Zhang, Shaoqing Ren, and Jian Sun.
\newblock Deep residual learning for image recognition.
\newblock In \emph{Proceedings of the IEEE Conference on Computer Vision and
  Pattern Recognition}, 2016.

\bibitem[He et~al.(2017)He, Gkioxari, Doll{\'a}r, and Girshick]{MaskRCNN}
Kaiming He, Georgia Gkioxari, Piotr Doll{\'a}r, and Ross Girshick.
\newblock Mask r-cnn.
\newblock In \emph{Proceedings of the IEEE international conference on computer
  vision}, pp.\  2961--2969, 2017.

\bibitem[Heusel et~al.(2017)Heusel, Ramsauer, Unterthiner, Nessler, and
  Hochreiter]{FID}
Martin Heusel, Hubert Ramsauer, Thomas Unterthiner, Bernhard Nessler, and Sepp
  Hochreiter.
\newblock Gans trained by a two time-scale update rule converge to a local nash
  equilibrium.
\newblock In \emph{Proceedings of the Neural Information Processing Systems
  Conference}, 2017.

\bibitem[Hinton et~al.(2015)Hinton, Vinyals, Dean,
  et~al.]{KnowledgeDistillation}
Geoffrey Hinton, Oriol Vinyals, Jeff Dean, et~al.
\newblock Distilling the knowledge in a neural network.
\newblock \emph{arXiv preprint arXiv:1503.02531}, 2\penalty0 (7), 2015.

\bibitem[Hua et~al.(2020)Hua, Kohli, Uplavikar, Ravi, Gunaseelan, Orozco, and
  Li]{Holopix50k}
Yiwen Hua, Puneet Kohli, Pritish Uplavikar, Anand Ravi, Saravana Gunaseelan,
  Jason Orozco, and Edward Li.
\newblock Holopix50k: A large-scale in-the-wild stereo image dataset.
\newblock In \emph{CVPR Workshop on Computer Vision for Augmented and Virtual
  Reality, Seattle, WA, 2020.}, June 2020.

\bibitem[Karras et~al.(2019)Karras, Laine, and Aila]{StyleGAN}
Tero Karras, Samuli Laine, and Timo Aila.
\newblock A style-based generator architecture for generative adversarial
  networks.
\newblock In \emph{Proceedings of the IEEE Conference on Computer Vision and
  Pattern Recognition}, 2019.

\bibitem[Karras et~al.(2020{\natexlab{a}})Karras, Aittala, Hellsten, Laine,
  Lehtinen, and Aila]{StyleGAN2-ADA}
Tero Karras, Miika Aittala, Janne Hellsten, Samuli Laine, Jaakko Lehtinen, and
  Timo Aila.
\newblock Training generative adversarial networks with limited data.
\newblock \emph{arXiv preprint arXiv:2006.06676}, 2020{\natexlab{a}}.

\bibitem[Karras et~al.(2020{\natexlab{b}})Karras, Laine, Aittala, Hellsten,
  Lehtinen, and Aila]{StyleGAN2}
Tero Karras, Samuli Laine, Miika Aittala, Janne Hellsten, Jaakko Lehtinen, and
  Timo Aila.
\newblock Analyzing and improving the image quality of stylegan.
\newblock In \emph{Proceedings of the IEEE Conference on Computer Vision and
  Pattern Recognition}, 2020{\natexlab{b}}.

\bibitem[Karras et~al.(2021)Karras, Aittala, Laine, H{\"a}rk{\"o}nen, Hellsten,
  Lehtinen, and Aila]{AliasFreeGAN}
Tero Karras, Miika Aittala, Samuli Laine, Erik H{\"a}rk{\"o}nen, Janne
  Hellsten, Jaakko Lehtinen, and Timo Aila.
\newblock Alias-free generative adversarial networks.
\newblock \emph{arXiv preprint arXiv:2106.12423}, 2021.

\bibitem[Kim et~al.(2022)Kim, Seo, and Han]{InfoNeRF}
Mijeong Kim, Seonguk Seo, and Bohyung Han.
\newblock Infonerf: Ray entropy minimization for few-shot neural volume
  rendering.
\newblock In \emph{CVPR}, 2022.

\bibitem[Kim et~al.(2021)Kim, Oh, Kim, Cho, and Yun]{CompareKLandL2forKD}
Taehyeon Kim, Jaehoon Oh, NakYil Kim, Sangwook Cho, and Se-Young Yun.
\newblock Comparing kullback-leibler divergence and mean squared error loss in
  knowledge distillation.
\newblock \emph{arXiv preprint arXiv:2105.08919}, 2021.

\bibitem[Kingma \& Ba(2014)Kingma and Ba]{Adam}
Diederik~P Kingma and Jimmy Ba.
\newblock Adam: A method for stochastic optimization.
\newblock \emph{arXiv preprint arXiv:1412.6980}, 2014.

\bibitem[Kuang et~al.(2022)Kuang, Olszewski, Chai, Huang, Achlioptas, and
  Tulyakov]{NeROIC}
Zhengfei Kuang, Kyle Olszewski, Menglei Chai, Zeng Huang, Panos Achlioptas, and
  Sergey Tulyakov.
\newblock Neroic: Neural rendering of objects from online image collections.
\newblock In \emph{Special Interest Group on Computer Graphics and Interactive
  Techniques}, 2022.

\bibitem[Kumari et~al.(2022)Kumari, Zhang, Shechtman, and
  Zhu]{EnsemblingOffTheShelfModels}
Nupur Kumari, Richard Zhang, Eli Shechtman, and Jun-Yan Zhu.
\newblock Ensembling off-the-shelf models for gan training.
\newblock In \emph{Proceedings of the IEEE Conference on Computer Vision and
  Pattern Recognition}, 2022.

\bibitem[Lin et~al.(2021)Lin, Ma, Torralba, and Lucey]{BaRF}
Chen-Hsuan Lin, Wei-Chiu Ma, Antonio Torralba, and Simon Lucey.
\newblock Barf: Bundle-adjusting neural radiance fields.
\newblock In \emph{Proceedings of the IEEE International Conference on Computer
  Vision}, 2021.

\bibitem[Meng et~al.(2021)Meng, Chen, Luo, Wu, Su, Xu, He, and Yu]{GNeRF}
Quan Meng, Anpei Chen, Haimin Luo, Minye Wu, Hao Su, Lan Xu, Xuming He, and
  Jingyi Yu.
\newblock Gnerf: Gan-based neural radiance field without posed camera.
\newblock In \emph{Proceedings of the IEEE International Conference on Computer
  Vision}, 2021.

\bibitem[Mescheder et~al.(2018)Mescheder, Geiger, and Nowozin]{R1_reg}
Lars Mescheder, Andreas Geiger, and Sebastian Nowozin.
\newblock Which training methods for gans do actually converge?
\newblock In \emph{International conference on machine learning}, 2018.

\bibitem[Miangoleh et~al.(2021)Miangoleh, Dille, Mai, Paris, and Aksoy]{LeReS}
S~Mahdi~H Miangoleh, Sebastian Dille, Long Mai, Sylvain Paris, and Yagiz Aksoy.
\newblock Boosting monocular depth estimation models to high-resolution via
  content-adaptive multi-resolution merging.
\newblock In \emph{Proceedings of the IEEE/CVF Conference on Computer Vision
  and Pattern Recognition}, pp.\  9685--9694, 2021.

\bibitem[Mildenhall et~al.(2020)Mildenhall, Srinivasan, Tancik, Barron,
  Ramamoorthi, and Ng]{NeRF}
Ben Mildenhall, Pratul~P Srinivasan, Matthew Tancik, Jonathan~T Barron, Ravi
  Ramamoorthi, and Ren Ng.
\newblock Nerf: Representing scenes as neural radiance fields for view
  synthesis.
\newblock In \emph{European conference on computer vision}, 2020.

\bibitem[Mo et~al.(2020)Mo, Cho, and Shin]{FreezeD}
Sangwoo Mo, Minsu Cho, and Jinwoo Shin.
\newblock Freeze the discriminator: a simple baseline for fine-tuning gans.
\newblock \emph{arXiv preprint arXiv:2002.10964}, 2020.

\bibitem[Mokady et~al.(2022)Mokady, Yarom, Tov, Lang, Cohen-Or, Dekel, Irani,
  and Mosseri]{SDIP}
Ron Mokady, Michal Yarom, Omer Tov, Oran Lang, Daniel Cohen-Or, Tali Dekel,
  Michal Irani, and Inbar Mosseri.
\newblock Self-distilled stylegan: Towards generation from internet photos.
\newblock \emph{arXiv preprint arXiv:2202.12211}, 2022.

\bibitem[Niemeyer \& Geiger(2021{\natexlab{a}})Niemeyer and Geiger]{CAMPARI}
Michael Niemeyer and Andreas Geiger.
\newblock Campari: Camera-aware decomposed generative neural radiance fields.
\newblock In \emph{2021 International Conference on 3D Vision (3DV)},
  2021{\natexlab{a}}.

\bibitem[Niemeyer \& Geiger(2021{\natexlab{b}})Niemeyer and Geiger]{Giraffe}
Michael Niemeyer and Andreas Geiger.
\newblock Giraffe: Representing scenes as compositional generative neural
  feature fields.
\newblock In \emph{Proceedings of the IEEE Conference on Computer Vision and
  Pattern Recognition}, 2021{\natexlab{b}}.

\bibitem[Odena et~al.(2018)Odena, Buckman, Olsson, Brown, Olah, Raffel, and
  Goodfellow]{JacobianForGenerator}
Augustus Odena, Jacob Buckman, Catherine Olsson, Tom Brown, Christopher Olah,
  Colin Raffel, and Ian Goodfellow.
\newblock Is generator conditioning causally related to gan performance?
\newblock In \emph{International conference on machine learning}, 2018.

\bibitem[Or-El et~al.(2021)Or-El, Luo, Shan, Shechtman, Park, and
  Kemelmacher-Shlizerman]{StyleSDF}
Roy Or-El, Xuan Luo, Mengyi Shan, Eli Shechtman, Jeong~Joon Park, and Ira
  Kemelmacher-Shlizerman.
\newblock Style{SDF}: {H}igh-{R}esolution {3D}-{C}onsistent {I}mage and
  {G}eometry {G}eneration.
\newblock \emph{arXiv preprint arXiv:2112.11427}, 2021.

\bibitem[Parmar et~al.(2021)Parmar, Zhang, and Zhu]{FID_evaluation}
Gaurav Parmar, Richard Zhang, and Jun-Yan Zhu.
\newblock On buggy resizing libraries and surprising subtleties in fid
  calculation.
\newblock \emph{arXiv preprint arXiv:2104.11222}, 2021.

\bibitem[pmneila(2015)]{PyMCubes}
pmneila.
\newblock Pymcubes.
\newblock \emph{https://github.com/pmneila/PyMCubes}, 2015.

\bibitem[Qi et~al.(2017{\natexlab{a}})Qi, Su, Mo, and Guibas]{PointNet}
Charles~R Qi, Hao Su, Kaichun Mo, and Leonidas~J Guibas.
\newblock Pointnet: Deep learning on point sets for 3d classification and
  segmentation.
\newblock In \emph{Proceedings of the IEEE conference on computer vision and
  pattern recognition}, pp.\  652--660, 2017{\natexlab{a}}.

\bibitem[Qi et~al.(2017{\natexlab{b}})Qi, Yi, Su, and Guibas]{PointNet++}
Charles~Ruizhongtai Qi, Li~Yi, Hao Su, and Leonidas~J Guibas.
\newblock Pointnet++: Deep hierarchical feature learning on point sets in a
  metric space.
\newblock \emph{Advances in neural information processing systems}, 30,
  2017{\natexlab{b}}.

\bibitem[Radford et~al.(2021)Radford, Kim, Hallacy, Ramesh, Goh, Agarwal,
  Sastry, Askell, Mishkin, Clark, et~al.]{CLIP}
Alec Radford, Jong~Wook Kim, Chris Hallacy, Aditya Ramesh, Gabriel Goh,
  Sandhini Agarwal, Girish Sastry, Amanda Askell, Pamela Mishkin, Jack Clark,
  et~al.
\newblock Learning transferable visual models from natural language
  supervision.
\newblock In \emph{International Conference on Machine Learning}, pp.\
  8748--8763. PMLR, 2021.

\bibitem[Ramesh et~al.(2022)Ramesh, Dhariwal, Nichol, Chu, and Chen]{DALLE-2}
Aditya Ramesh, Prafulla Dhariwal, Alex Nichol, Casey Chu, and Mark Chen.
\newblock Hierarchical text-conditional image generation with clip latents.
\newblock \emph{arXiv preprint arXiv:2204.06125}, 2022.

\bibitem[Rombach et~al.(2022)Rombach, Blattmann, Lorenz, Esser, and Ommer]{LDM}
Robin Rombach, Andreas Blattmann, Dominik Lorenz, Patrick Esser, and Bj{\"o}rn
  Ommer.
\newblock High-resolution image synthesis with latent diffusion models.
\newblock In \emph{Proceedings of the IEEE/CVF Conference on Computer Vision
  and Pattern Recognition}, pp.\  10684--10695, 2022.

\bibitem[Salimans et~al.(2016)Salimans, Goodfellow, Zaremba, Cheung, Radford,
  and Chen]{InceptionScore}
Tim Salimans, Ian Goodfellow, Wojciech Zaremba, Vicki Cheung, Alec Radford, and
  Xi~Chen.
\newblock Improved techniques for training gans.
\newblock In \emph{Proceedings of the Neural Information Processing Systems
  Conference}, 2016.

\bibitem[Sauer et~al.(2021)Sauer, Chitta, M{\"u}ller, and
  Geiger]{ProjectedGANs}
Axel Sauer, Kashyap Chitta, Jens M{\"u}ller, and Andreas Geiger.
\newblock Projected gans converge faster.
\newblock \emph{Advances in Neural Information Processing Systems},
  34:\penalty0 17480--17492, 2021.

\bibitem[Sauer et~al.(2022)Sauer, Schwarz, and Geiger]{StyleGAN-XL}
Axel Sauer, Katja Schwarz, and Andreas Geiger.
\newblock Stylegan-xl: Scaling stylegan to large diverse datasets.
\newblock In \emph{ACM SIGGRAPH 2022 Conference Proceedings}, 2022.

\bibitem[Sch\"{o}nberger \& Frahm(2016)Sch\"{o}nberger and Frahm]{SFM}
Johannes~Lutz Sch\"{o}nberger and Jan-Michael Frahm.
\newblock Structure-from-motion revisited.
\newblock In \emph{Proceedings of the IEEE Conference on Computer Vision and
  Pattern Recognition}, 2016.

\bibitem[Sch\"{o}nberger et~al.(2016)Sch\"{o}nberger, Zheng, Pollefeys, and
  Frahm]{schoenberger2016mvs}
Johannes~Lutz Sch\"{o}nberger, Enliang Zheng, Marc Pollefeys, and Jan-Michael
  Frahm.
\newblock Pixelwise view selection for unstructured multi-view stereo.
\newblock In \emph{Proceedings of the European Conference on Computer Vision},
  2016.

\bibitem[Schwarz et~al.(2020)Schwarz, Liao, Niemeyer, and Geiger]{GRAF}
Katja Schwarz, Yiyi Liao, Michael Niemeyer, and Andreas Geiger.
\newblock Graf: Generative radiance fields for 3d-aware image synthesis.
\newblock In \emph{Proceedings of the Neural Information Processing Systems
  Conference}, 2020.

\bibitem[Schwarz et~al.(2022)Schwarz, Sauer, Niemeyer, Liao, and
  Geiger]{VoxGRAF}
Katja Schwarz, Axel Sauer, Michael Niemeyer, Yiyi Liao, and Andreas Geiger.
\newblock Voxgraf: Fast 3d-aware image synthesis with sparse voxel grids.
\newblock \emph{ARXIV}, 2022.

\bibitem[Shi et~al.(2022)Shi, Shen, Zhu, Yeung, and Chen]{DepthGAN}
Zifan Shi, Yujun Shen, Jiapeng Zhu, Dit-Yan Yeung, and Qifeng Chen.
\newblock 3d-aware indoor scene synthesis with depth priors.
\newblock 2022.

\bibitem[Shih et~al.(2020)Shih, Su, Kopf, and Huang]{3DPhoto}
Meng-Li Shih, Shih-Yang Su, Johannes Kopf, and Jia-Bin Huang.
\newblock 3d photography using context-aware layered depth inpainting.
\newblock In \emph{IEEE Conference on Computer Vision and Pattern Recognition
  (CVPR)}, 2020.

\bibitem[Shu et~al.(2019)Shu, Park, and Kwon]{FPD}
Dong~Wook Shu, Sung~Woo Park, and Junseok Kwon.
\newblock 3d point cloud generative adversarial network based on tree
  structured graph convolutions.
\newblock In \emph{Proceedings of the IEEE/CVF international conference on
  computer vision}, pp.\  3859--3868, 2019.

\bibitem[Skorokhodov et~al.(2022)Skorokhodov, Tulyakov, Wang, and
  Wonka]{EpiGRAF}
Ivan Skorokhodov, Sergey Tulyakov, Yiqun Wang, and Peter Wonka.
\newblock Epigraf: Rethinking training of 3d gans.
\newblock In \emph{Proceedings of the Neural Information Processing Systems
  Conference}, 2022.

\bibitem[Sun et~al.(2022)Sun, Wang, Shi, Wang, Wang, and Liu]{IDE-3D}
Jingxiang Sun, Xuan Wang, Yichun Shi, Lizhen Wang, Jue Wang, and Yebin Liu.
\newblock Ide-3d: Interactive disentangled editing for high-resolution 3d-aware
  portrait synthesis.
\newblock \emph{arXiv preprint arXiv:2205.15517}, 2022.

\bibitem[Tan \& Le(2019)Tan and Le]{EfficientNet}
Mingxing Tan and Quoc Le.
\newblock Efficientnet: Rethinking model scaling for convolutional neural
  networks.
\newblock In \emph{International Conference on Machine Learning}, 2019.

\bibitem[Wang et~al.(2022)Wang, Chai, He, Chen, and Liao]{CLIP-NeRF}
Can Wang, Menglei Chai, Mingming He, Dongdong Chen, and Jing Liao.
\newblock Clip-nerf: Text-and-image driven manipulation of neural radiance
  fields.
\newblock In \emph{Proceedings of the IEEE Conference on Computer Vision and
  Pattern Recognition}, 2022.

\bibitem[Wang et~al.(2021)Wang, Wu, Xie, Chen, and Prisacariu]{NeRF--}
Zirui Wang, Shangzhe Wu, Weidi Xie, Min Chen, and Victor~Adrian Prisacariu.
\newblock Ne{RF}$--$: Neural radiance fields without known camera parameters.
\newblock \emph{arXiv preprint arXiv:2102.07064}, 2021.

\bibitem[Xian et~al.(2020)Xian, Zhang, Wang, Mai, Lin, and Cao]{HRWSI}
Ke~Xian, Jianming Zhang, Oliver Wang, Long Mai, Zhe Lin, and Zhiguo Cao.
\newblock Structure-guided ranking loss for single image depth prediction.
\newblock In \emph{The IEEE/CVF Conference on Computer Vision and Pattern
  Recognition (CVPR)}, June 2020.

\bibitem[Xu et~al.(2021)Xu, Peng, Yang, Shen, and Zhou]{VolumeGAN}
Yinghao Xu, Sida Peng, Ceyuan Yang, Yujun Shen, and Bolei Zhou.
\newblock 3d-aware image synthesis via learning structural and textural
  representations.
\newblock \emph{arXiv preprint arXiv:2112.10759}, 2021.

\bibitem[Xue et~al.(2022)Xue, Li, Singh, and Lee]{GIRAFFE-HD}
Yang Xue, Yuheng Li, Krishna~Kumar Singh, and Yong~Jae Lee.
\newblock Giraffe hd: A high-resolution 3d-aware generative model.
\newblock \emph{arXiv preprint arXiv:2203.14954}, 2022.

\bibitem[Yan(2019)]{Pytorch_Pointnet_Pointnet2}
Xu~Yan.
\newblock Pointnet/pointnet++ pytorch.
\newblock \emph{https://github.com/yanx27/Pointnet\_Pointnet2\_pytorch}, 2019.

\bibitem[Yin et~al.(2021)Yin, Liu, and Shen]{DiverseDepth}
Wei Yin, Yifan Liu, and Chunhua Shen.
\newblock Virtual normal: Enforcing geometric constraints for accurate and
  robust depth prediction.
\newblock \emph{IEEE Transactions on Pattern Analysis and Machine Intelligence
  (TPAMI)}, 2021.

\bibitem[Yu et~al.(2015)Yu, Seff, Zhang, Song, Funkhouser, and Xiao]{LSUN}
Fisher Yu, Ari Seff, Yinda Zhang, Shuran Song, Thomas Funkhouser, and Jianxiong
  Xiao.
\newblock Lsun: Construction of a large-scale image dataset using deep learning
  with humans in the loop.
\newblock \emph{arXiv preprint arXiv:1506.03365}, 2015.

\bibitem[Yu et~al.(2022)Yu, Peng, Niemeyer, Sattler, and Geiger]{MonoSDF}
Zehao Yu, Songyou Peng, Michael Niemeyer, Torsten Sattler, and Andreas Geiger.
\newblock Monosdf: Exploring monocular geometric cues for neural implicit
  surface reconstruction.
\newblock \emph{arXiv preprint arXiv:2206.00665}, 2022.

\bibitem[Zamir et~al.(2018)Zamir, Sax, Shen, Guibas, Malik, and
  Savarese]{Taskonomy}
Amir~R Zamir, Alexander Sax, William Shen, Leonidas~J Guibas, Jitendra Malik,
  and Silvio Savarese.
\newblock Taskonomy: Disentangling task transfer learning.
\newblock In \emph{Proceedings of the IEEE conference on computer vision and
  pattern recognition}, pp.\  3712--3722, 2018.

\bibitem[Zhang et~al.(2022)Zhang, Zheng, Gao, Zhang, Pan, and Yang]{MVCGAN}
Xuanmeng Zhang, Zhedong Zheng, Daiheng Gao, Bang Zhang, Pan Pan, and Yi~Yang.
\newblock Multi-view consistent generative adversarial networks for 3d-aware
  image synthesis.
\newblock In \emph{Proceedings of the IEEE Conference on Computer Vision and
  Pattern Recognition}, 2022.

\bibitem[Zhao et~al.(2022)Zhao, Ma, Güera, Ren, Schwing, and Colburn]{GMPI}
Xiaoming Zhao, Fangchang Ma, David Güera, Zhile Ren, Alexander~G. Schwing, and
  Alex Colburn.
\newblock Generative multiplane images: Making a 2d gan 3d-aware.
\newblock In \emph{Proc. ECCV}, 2022.

\bibitem[Zhou et~al.(2021)Zhou, Xie, Ni, and Tian]{CIPS-3D}
Peng Zhou, Lingxi Xie, Bingbing Ni, and Qi~Tian.
\newblock Cips-3d: A 3d-aware generator of gans based on
  conditionally-independent pixel synthesis.
\newblock \emph{arXiv preprint arXiv:2110.09788}, 2021.

\end{thebibliography}
\bibliographystyle{iclr2023_conference}

\clearpage
\appendix
\section{Limitations}\label{ap:limitations}


\textbf{Lower \textbf{visual} quality compared to 2D generators}.
Despite providing a more reasonable representation of the underlining scene, 3D generators still have a lower visual quality compared to 2D generators.
Closing this gap is essential for a wide adaptation of 3D generators.

\textbf{Background sticking}. One common 3D artifact of {\modelname} is gluing of the foreground and the background. In other words our model predicts a single shape for both and there is no clear separation between the two. One potential cause of this artifact is the dataset bias, where most of the photos are frontal, thus it is not beneficial for the model to explore backward views. Another reason is the bias of tri-planes toward flat geometry. However, all our attempts to replace tri-planes with an MLP-based NeRF led to much worse results (see Appx~\ref{ap:failed-experiments}).
Inventing a different efficient parametrization may be an important direction.

\textbf{No reliable quantitative measure of generated geometry}. In this work we introduce {\metricfullname} as a proxy metric for evaluating the quality of the underlying geometry. However it can capture only a single failure case, specific for generators based on tri-planes: flatness of geometry. Devising a reasonable metric applicable to a variety of scenarios could significantly speed up the progress in this area.

\begin{figure}[h!]
\centering
\includegraphics[width=0.7\textwidth]{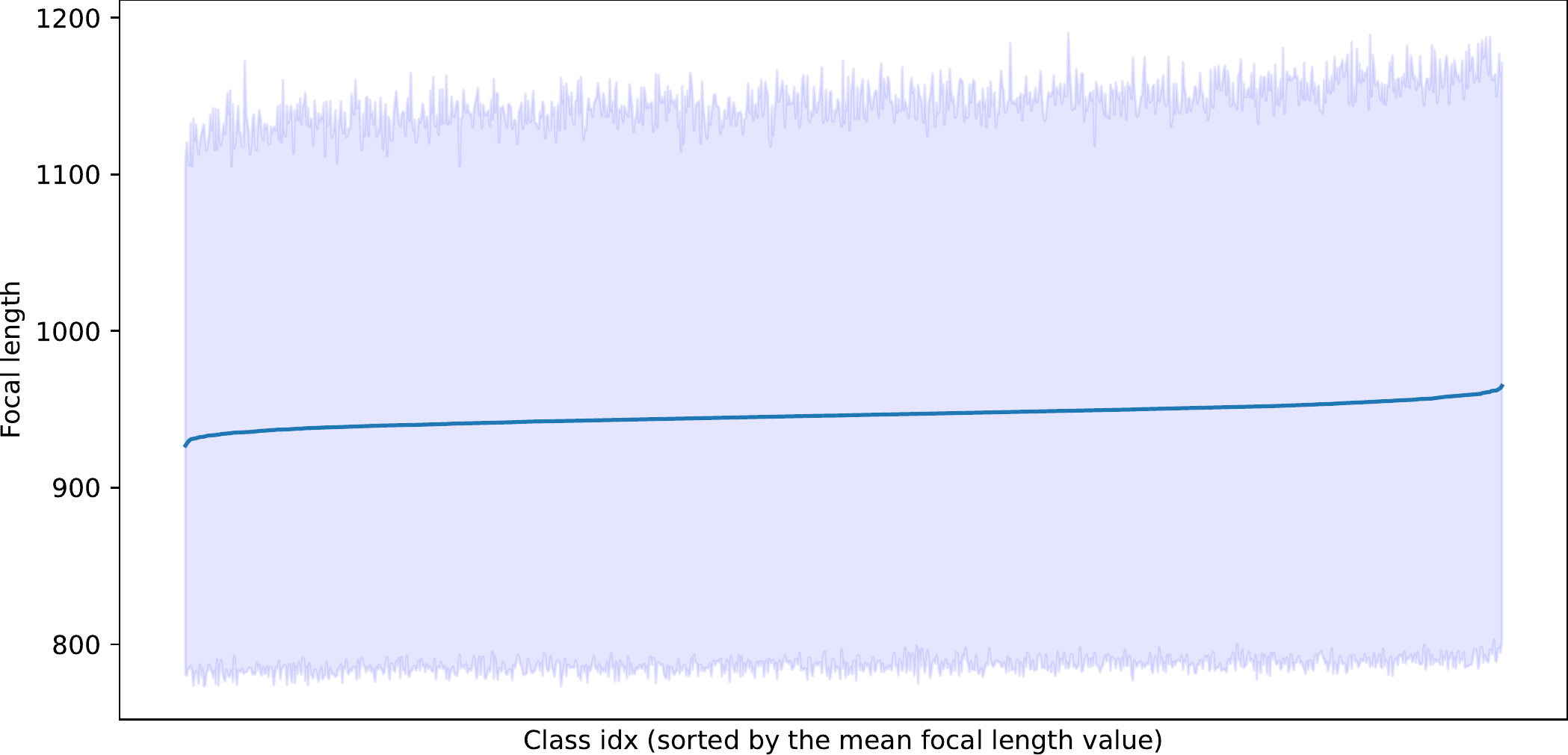}
\caption{Focal length distribution on ImageNet $256^2$ learned by $\C$. The blue solid line is the mean values, while lower/upper curves are 0.05 and 0.95 quantiles, respectively.}
\label{fig:focal-length-dist}
\end{figure}

\textbf{Camera generator $\C$ does not learn fine-grained control}. 
While our camera generator is conditioned on the class label $\bm c$, and, in theory, it should be able to perform fine-grained control over the class focal length distributions (which is natural since landscape panoramas and close-up view of a coffee mug typically have different focal lengths), it does not do this, as shown in \figref{fig:focal-length-dist}.
We attribute this problem to the implicit bias of $\G$ to produce large-FoV images due to tri-planes parametrization.
Tri-planes define a limited volume box in space, and close-up renderings with large focal length would utilize fewer tri-plane features, hence using less generator's capacity.
This is why \modelname\ attempts to perform modeling with larger field-of-view values.

\section{Implementation details}\label{ap:implementation-details}

\begin{table}
\centering
\caption{Module names glossary.}
\begin{tabular}{cc}
 \toprule
 $\G$ & Generator \\
 $\M$ & Mapping network \\
 $\Ss$ & Synthesis network \\
 $\T$ & Tri-plane decoder \\
 $\C$ & Camera generator \\
 $\A$ & Depth adaptor \\
 $\D$ & Discriminator  \\
 $\E$ & Depth Estimator \\
 $\F$ & Feature Extractor \\
 \bottomrule
\end{tabular}
\label{tab:notation}
\end{table}
This section provides additional architectural and training details.
Also, note that we release the source code.
Our generator $\G$ consists of $\M$, $\Ss$ and $\T$.
Mapping network $\M$ takes noise $\bm z \in \mathbb{R}^{512}$ and class label $c \in {0, ..., K-1}$, where K is the number of classes, and produces the style code $\bm w \in \R^{512}$.
Similar to StyleGAN2-ADA~\cite{StyleGAN2-ADA} and EpiGRAF~\cite{EpiGRAF}, $\M$ is 2-layer MLP network with LeakyReLU activations and 512 neurons in each layer.
Synthesis network is a decoder network same as StyleGAN2~\cite{StyleGAN2}, except that it produce tri-plane feature features $\bm p = (\bm p^{xy}, \bm p^{yz}, \bm p^{xz}) \in \R^{3 \times (512 \times 512 \times 32)}$.
A feature vector $\bm f_{xyz} \in \R^{32}$ located $(x, y, z) \in \R^3$ is computed by projecting the coordinate back to the tri-plane representation, followed by bi-linearly interpolating the nearby features and averaging the features from different planes.
Following EpiGRAF~\citep{EpiGRAF}, tri-plane decoder is a two-layer MLP with Leaky-ReLU activations and 64 neurons in the hidden layer, that takes a tri-plane feature $\bm f_{xyz}$ as input and produced the color and density $(\text{RGB}, \sigma)$ in that point.
We use same volume rendering procedure as EpiGRAF~\citep{EpiGRAF}.
We also found that increasing the half-life of the exponential moving average 
for our $\G$ improves both FID and Inception Score.
We observed this by noticing that the samples change too rapidly over the course of training.
In practice, we log the samples every 400k seen images and noticed that the generator could completely change the global structure of a sample during that period.

\begin{figure}
\centering
\begin{subfigure}[b]{0.55\textwidth}
    \centering
    \includegraphics[width=\linewidth]{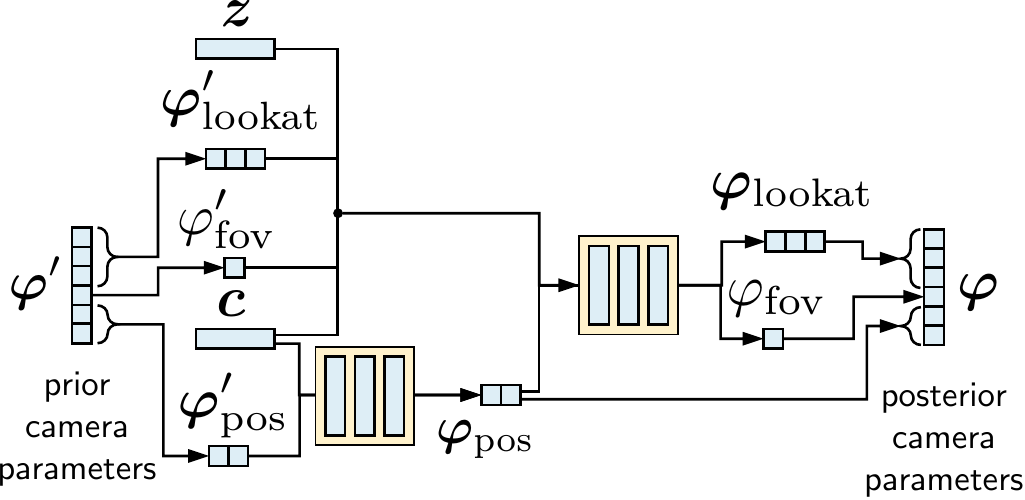}
    \caption{\textbf{Camera generator $\C$.} We condition $\C$ on class labels $\bm c$ when generating the camera position $\hat{\bm \varphi}$ since it might be different for different classes. And we condition it on $\bm z$ when generating the look-at position and field-of-view since it might depend on the object shape (e.g., there is a higher probability to synthesize a close-up view of a dog's snout rather than its tail). Each MLP consists on 3 layers with Softplus non-linearities. }
    \label{fig:camera-generator:architecture}
\end{subfigure}
\hfill
\begin{subfigure}[b]{0.42\textwidth}
    \centering
    \includegraphics[width=\linewidth]{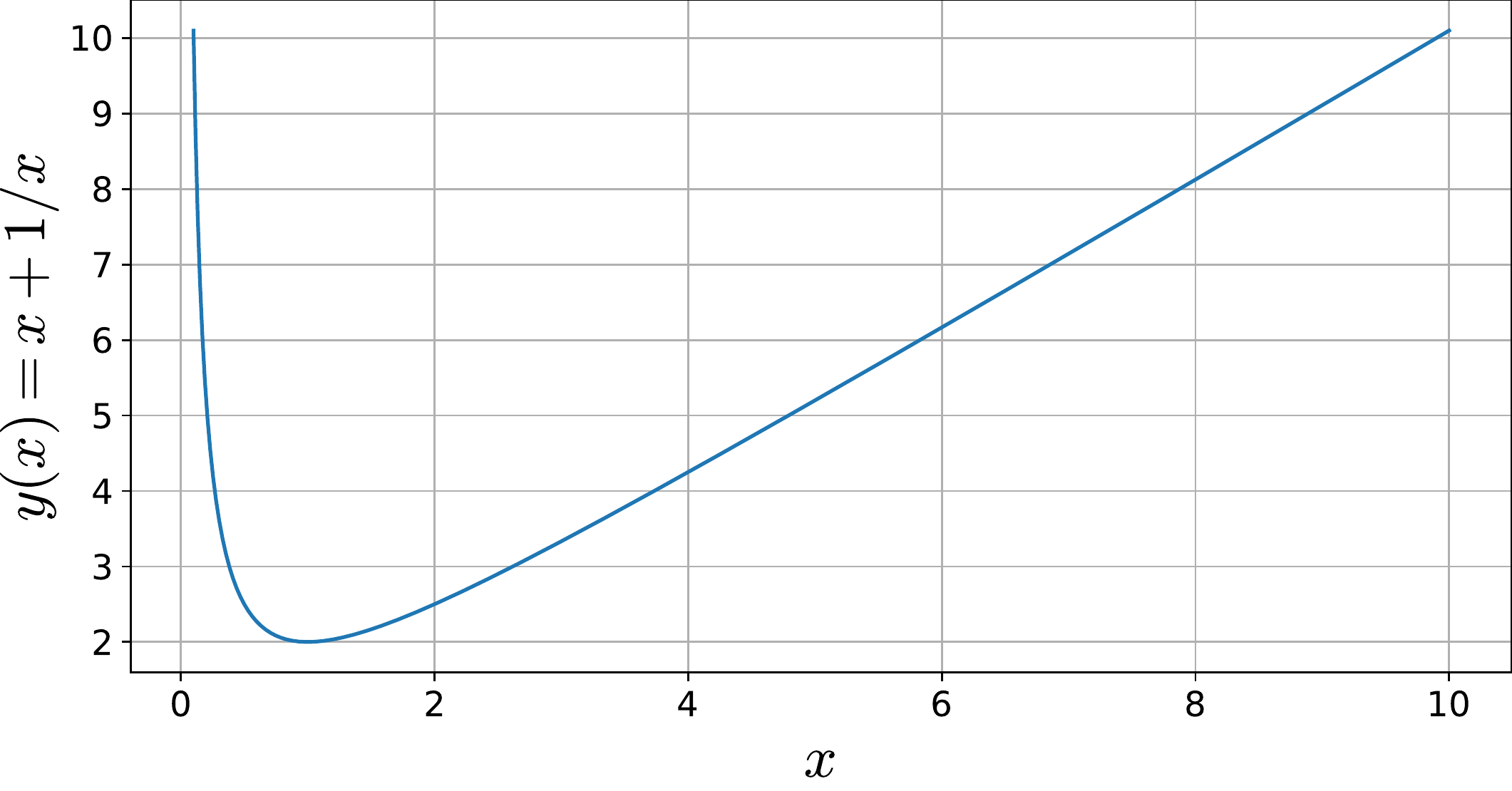}
    \caption{\textbf{Camera gradient penalty}. We structure the regularization term for $\C$ as a function $\mathcal{L}(|g|) = |g| + 1/|g|$ (see \eqref{eq:camera-gen-grad-penalty}), and this function is visualized above. This allows to prevent the collapse of $\C$ into delta distribution by it either producing constant values or very large/small values (which become constant after the sigmoid normalization).}
    \label{fig:camera-generator:reg}
\end{subfigure}
\caption{Camera generator architecture and visualizing its corresponding regularization term.}
\label{fig:camera-generator}
\end{figure}

\textbf{Camera generator}.
Camera generator consists of linear layers with SoftPlus activations, it architecture is depicted in \figref{fig:camera-generator}. We found it crucial to use SoftPlus activation and not LeakyReLU, since optimization of {\gradpen} for non-smooth functions is unstable for small learning rates (smaller 0.02).
Our gradient penalty minimizes the function $\mathcal{L} = |g| + 1/|g|$, where $|g|$ is the input/output scalar derivative of the camera generator $\C$.
The motivation is to prevent the collapse of $\C$ into delta distribution and the intuition is the following.
$\C$ can collapse into delta distribution in two ways: 1) by starting to produce the constant output for all the input values (this is being prevented by the first term $|g|$ and 2) by starting producing $\pm\infty$ for all the inputs, which are at the end converted to constants since we apply sigmoid normalization on top of its outputs to normalize them into a proper range (e.g., pitch is bounded in (0, $\pi$)) --- this, in turn, is prevented by the second term $1/|g|$. The minimum value of this function is 2 (which is achieved when the gradient norm is constant and equals to 1), and the function itself is visualized in \figref{fig:camera-generator:reg}.

\textbf{Adversarial depth supervision}. For depth adaptor, we use a three layer convolutional neural network with 5$\times$5 kernel sizes (since it is shallow, we increase its receptive field by increasing the kernel size) with LeakyReLU activations and 64 filters in each layer.
Additionally we use one shared convolutional layer that converts $64\times h \times w$ features to the depth maps.
We use the same architecture for $\D$ as EpiGRAF~\citep{EpiGRAF}, but additionally concatenate a 1-channel depth to the 3-channel RGB input.
Finally $\E$ and $\F$ is pretrainined LeReS~\citep{LeReS} and ResNet50~\citep{ResNet} networks without any modifications.
We used the \texttt{timm} library to extract the features for real images.

Similarly to EpiGRAF~\citep{EpiGRAF} we set the $\mathcal{R}_1$ regularization~\citep{R1_reg} term $\lambda_r=0.1$ and knowledge distillation term $\lambda_\text{dist} = 1$.
$\mathcal{R}_1$ regularization helps to stabilize GAN training and is formulated as a gradient penalty on top of the discriminator's real inputs:
\begin{equation}\label{eq:r1-reg}
\mathcal{R}_1 = \frac{1}{2}\| \nabla_{\bm x}\D(\bm{x}) \|_2^2 \longrightarrow \min_{\D}.
\end{equation}
The (hand-wavy) intuition is that it makes the discriminator surface more flat in the vicinity of real points, making it easier for the generator to reach them.
We train all the models with Adam optimizer~\citep{Adam} using the learning rate of $2e{\text -}3$ and $\beta_1 = 0.0, \beta_2 = 0.99$. Following EpiGRAF~\citep{EpiGRAF}, our model uses patch-wise training with $64\times64$-resolution patches and uses their proposed $\beta$ scale sampling strategy without any modifications. 
We use the batch size of 64 in all the experiments, since in early experiments we didn't find any improvements from using a large batch size neither for our model nor for StyleGAN2, as observed by \cite{BigGAN} and \cite{StyleGAN-XL}.




As being said in Section~\ref{sec:experiments}, we use the instance selection technique by \cite{InstanceSelectionForGANs} to remove image outliers from the datasets since they might negatively affect the geometry.
This procedure works by first extracting a 2,048-dimensional feature vector for each image, then fitting a multivariate gaussian distribution on the obtained dataset and removing the images, which features have low probability density.
For SDIP Dogs and LSUN Horses, we fit a multi-variate gaussian distribution for the whole dataset.
For ImageNet, we fit a separate model for each class with additional diagonal regularization for covariance, which is needed due to its singularity: feature vector has more dimensions than the number of images in a class.
We refer to \citep{InstanceSelectionForGANs} for additional details.

%


Also note that we will release the source code and the checkpoints, which would additionally convey all other implementation details.

\section{Knowledge distillation}\label{ap:knowledge-distillation}

In this section we compare our knowledge distillation strategy with projected $\D$~\citep{ProjectedGANs, StyleGAN-XL}, another popular technique of utilizing existing classification models for GAN training. Note that projected $\D$ relies on pretrained EfficientNet~\citep{EfficientNet}, it freezes most of the weights and adds some random convolutions layers that project intermediate outputs of EfficientNet to the features, that will be later processed by discriminator. Note that this strategy relies on specific architecture of Discriminator $\D$, thus in order to adapt it for {\modelname} we have to rely on two discriminators. First one is our disciminator $\D$ with additional depth input and conditioned on patch parameters and second one is projected $\D$. In Tab~\ref{tab:knowledge-distill} we show comparison between projected $\D$ and our knowledge distillation strategy on ImageNet~\citep{ImageNet} and \dogsf~\cite{SDIP} datasets. First we compare these strategies for 2D generator architectures: StyleGAN3-t (large)~\citep{AliasFreeGAN} and StyleGAN2~\citep{StyleGAN2-ADA}.
It can be observed that projected $\D$ is not generic and it results heavily varies depending on generator architecture. Moreover training cost of projective $\D$ is higher. Finally we test this strategy for our 3D generator, and again we can observe that this technique lead to inferior results in both visual quality and training cost.

\begin{table*}[h!]
\caption{Comparing our knowledge transfer strategy with the one from StyleGAN-XL. }
\label{tab:knowledge-distill}
\centering
\resizebox{1.0\linewidth}{!}{
\begin{tabular}{lccccc}
\toprule
\multirow{2}{*}{Method} & \multicolumn{2}{c}{ImageNet $128^2$ @ 10M} & \multicolumn{2}{c}{SDIP Dogs$_\text{40k}$ $256^2$ @ 5M} & \multirowcell{2}{Restricts $\D$'s \\ architecture?} \\
& FID$\downarrow$ & Training cost $\downarrow$ & FID $\downarrow$ & Training cost $\downarrow$ & \\
\midrule
StyleGAN3-t (large) & 28.1 & 11.9 & 6.42 & 7.3 & No \\
~- with Projected D & 22.8 & 11.6 & 22.6 & 8.1  & Yes \\
~- with Knowledge Distillation & 16.3 & 11.9 & 2.47 & 7.3  & No \\
\midrule
StyleGAN2 & 33.7 & 1.81 & 9.79 & 1.3 & No \\
~- with Projected D & 160.5 & 3.83 & 4.10 & 2.1 & Yes \\
~- with Knowledge Distillation with $\Ltwo$ & 20.75 & 1.83 & 2.08 & 1.4 & No \\
~- with Knowledge Distillation with KL & 25.92 & 1.83 & 2.54 & 1.4 & No \\
\midrule
\modelname & 53.6 & 6.33 & 21.3 & 2.6  & No \\
~- with Projected D & 105.9 & 8.0 & 10.5 & 4.2 & Yes \\
~- with Knowledge Distillation & 27.8 & 6.83 & 4.51 & 2.6 & No \\
\bottomrule
\end{tabular}
}
\end{table*}

\begin{figure}
\centering
\includegraphics[width=0.8\textwidth]{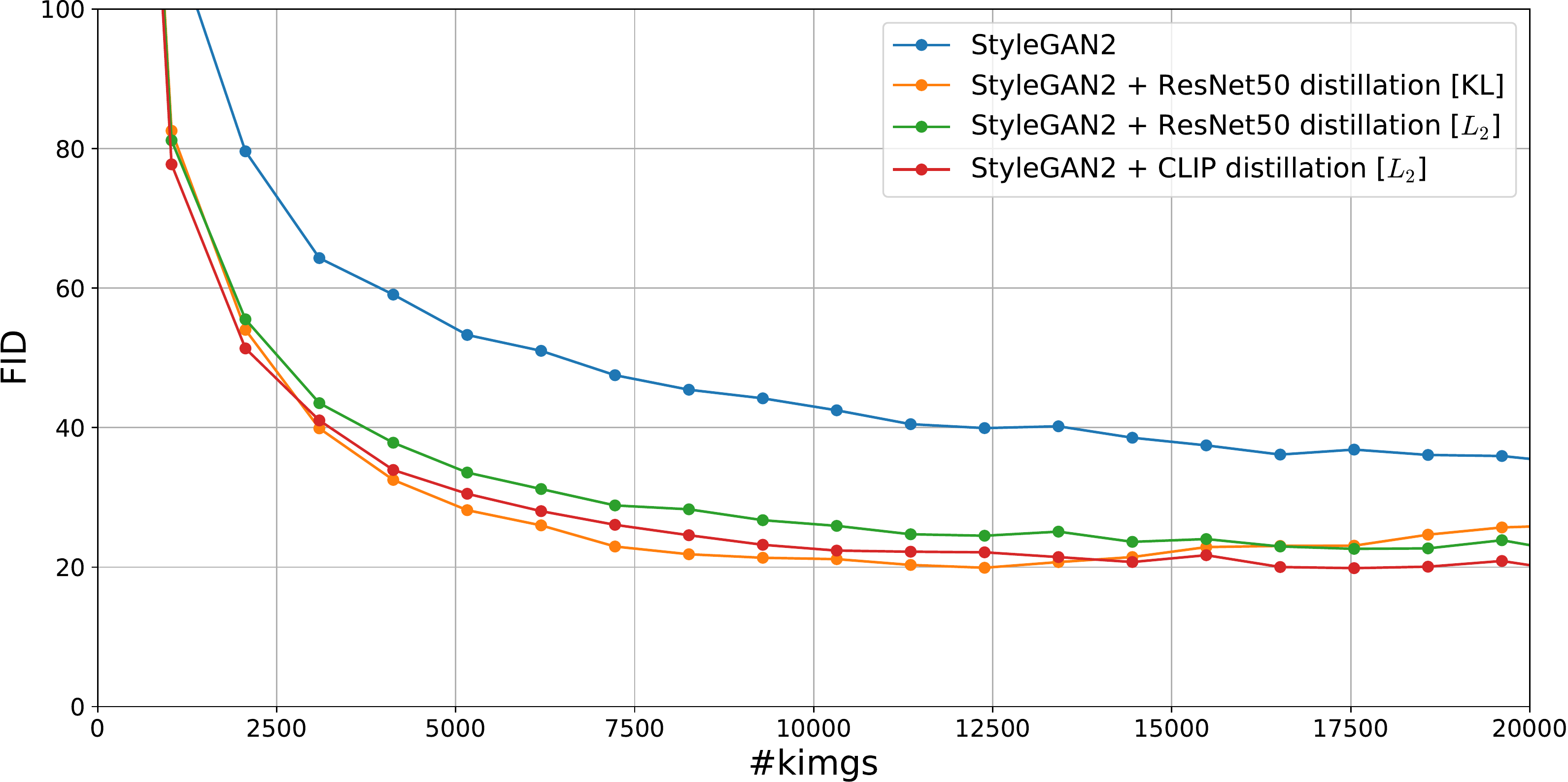}
\caption{\textbf{Exploring different knowledge distillation objectives}. Each model is trained for conditional ImageNet generation on the $128^2$ resolution with all other hyperparameters being the same.}
\label{fig:distillation}
\end{figure}

Typically, knowledge distillation in classifiers is performed using Kullback-Leibler divergence on top of logits~\citep{KnowledgeDistillation, CompareKLandL2forKD}, rather than $\mathcal{L}_2$ distance on top of hidden activations, as in our case.
There are two design reasons of why we use the $\Ltwo$: 1) it is more generalizable and one can transfer knowledge from other models with the same design, like CLIP~\citep{CLIP} or Mask R-CNN~\citep{MaskRCNN}; 2) our discriminator's task shouldn't be able to perform classification, that is why guiding it with the KL classification loss is not natural.
However, in \figref{fig:distillation}, we provide additional exploration with other knowledge distillation objectives on top of StyleGAN2~\citep{StyleGAN2} trained for conditional ImageNet generation on the $128^2$ resolution.
One can observe that KL and $\Ltwo$ objectives perform approximately the same.
Also, guiding by CLIP underperforms to ResNet guidance initially, but starts to outperform after more training is performed: we hypothesize that the reason is that it is a more general model.
We do not use CLIP guidance for our generators in this paper so not to give it an unfair advantage compared to other models.

\section{Failed experiments}\label{ap:failed-experiments}
In this section we describe several ideas that have been tried in this project, which however did not work for us.

\textbf{Gradient Penalty for Depth Adaptor}. To prevent depth adaptor $\A$ from completely faking the depth and forcing it to still provide a useful signal to the generator, one could employ a gradient penalty that will force gradient of $\A$ close to one. We tried it in two setups: window-to-pixel gradient and pixel-to-pixel gradient. Even with the large weight for this loss and shallow adaptor with $k=3$ kernel size, it didn't enforce good geometry

\textbf{{\modelname} with Projected $\D$}. Before devoting to knowledge distillation, we spend significant resources trying to incorporate projected $\D$ in our setup. For double discriminators setting, see Appx~\ref{ap:knowledge-distillation}, we tested: different weighting strategies, less frequent updates for projected $\D$, enabling projected $\D$ after some geometry was already learned. We also tested single projected $\D$, where we learned additional depth encoder that can consume our depth inputs. In all these experiments learned geometry was significantly inferior to the setting without projected $\D$. Most of the time geometry becomes completely flat.

\textbf{MLPs instead of tri-planes}. We noticed that tri-planes are extremely biased towards flat generation. Therefore one natural idea would be to utilize different representations, one possible candidate is MLPs~\cite{piGAN}. However our experiments with MLPs always lead to significantly inferior visual quality. Our hypothesis is that they just don't have enough capacity to model large scale datasets.

\textbf{Few-shot NeRF regularization}. We also try to improve learned geometry by incorporating ray entropy minimization loss from InfoNeRF~\cite{InfoNeRF} with different weights. But the model always diverges. We hypothesize that this is because it is significantly more complicated to find proper geometry, if the model is stuck in local minimum where the object densities are very sharp.

\textbf{Normal supervision}. Another prominent idea was utilizing normal supervision, since generic networks for normals~\cite{Omnidata} are also widely available and provide good results on the arbitrary data. Since computing normals requires gradient of density $\sigma$ with respect to $(x, y, z)$ and pytorch does not provide second order derivatives for \texttt{F.grid\_sample}~\footnote{https://github.com/pytorch/pytorch/issues/34704}, we developed a custom CUDA kernel that computes second derivative for \texttt{F.grid\_sample}. Unfortunately we found that normals supervision is inferior to depth and it is much easier for the generator $\G$ to fake normals.

\textbf{Preventing $\C$ collapse via variance/entropy/moments regularization}. Before arriving to our current form we had been experimenting with other forms of regularization: maximizing variance with some small loss coefficient, or entropy (in the assumption that the posterior distribution is gaussion), or additionally pushing mean/skewness/kurtosis to the one of the gaussian distribution. But each time, the generator was finding the ways to ``cheat'' the regularization and managing to produce either a delta distribution or a mixture of delta distributions (it very “likes” doing so since, it is able to completely flat images and cheat the geometry in this regime).

\section{Additional samples}\label{ap:additional-samples}

Since it is much easier to visualize the samples from NeRF-based generators as videos rather than RGB images, we provide all the additional visualizations on \projecthref.

\begin{figure}
    \centering
    \includegraphics[width=\textwidth]{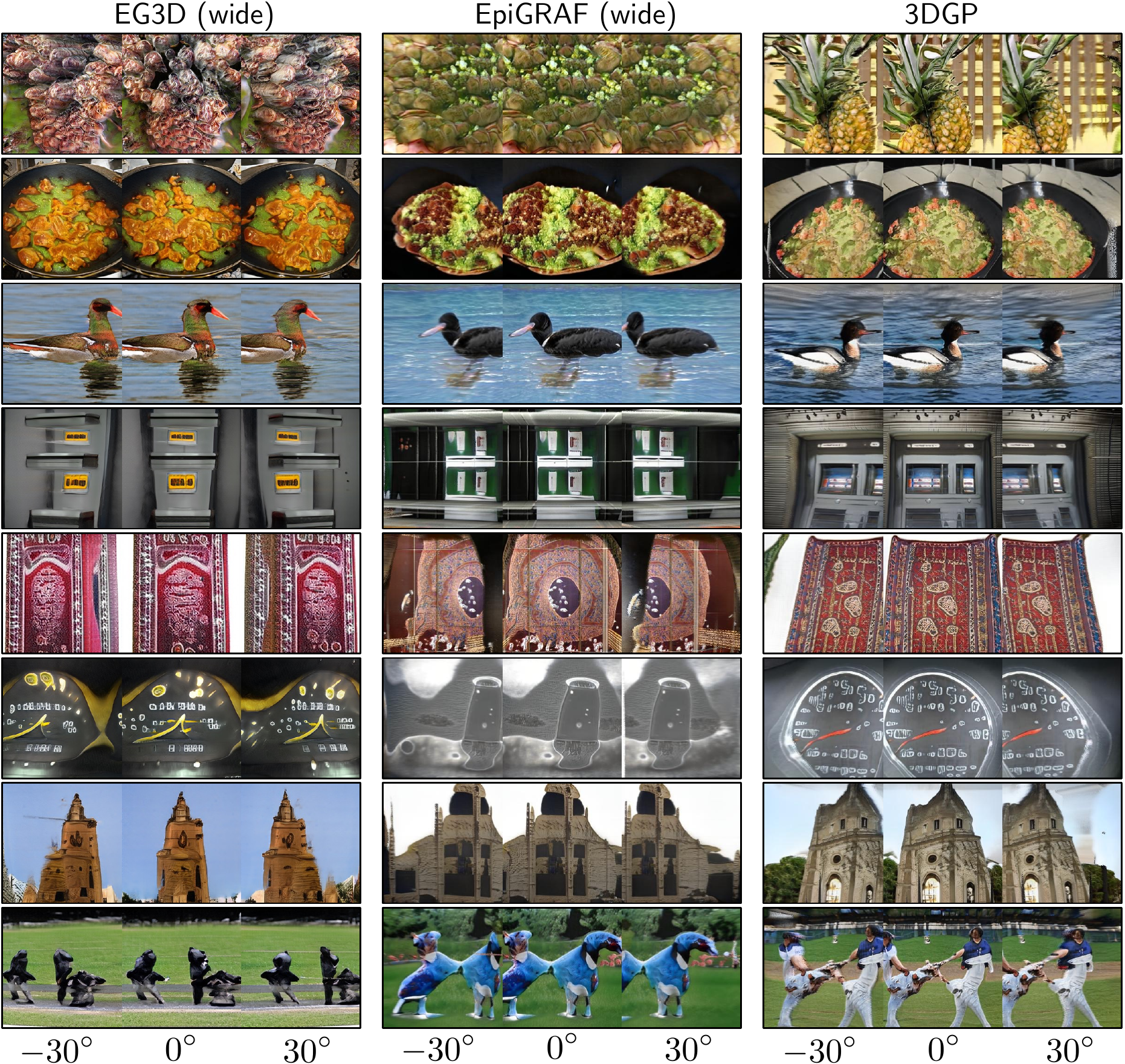}
    \caption{\emph{Random} samples for \emph{random} classes on ImageNet $256^2$ (random seed of 1 for the first ImageNet classes).}
    \label{fig:imagenet-samples}
\end{figure}

\begin{figure}
    \centering
    \includegraphics[width=0.8\textwidth]{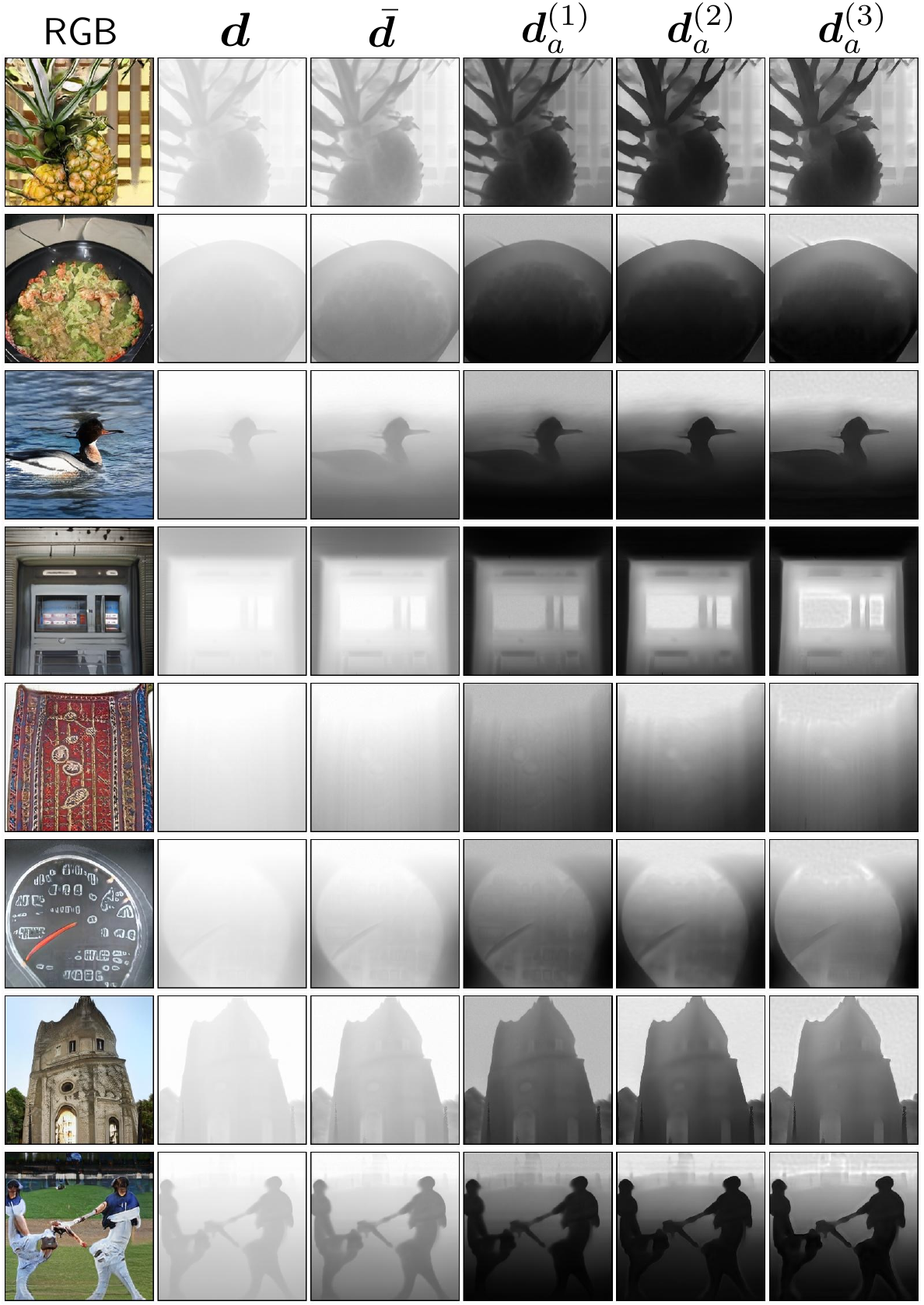}
    \caption{Depth maps produced by \modelname.}
    \label{fig:fake-depth}
\end{figure}

\begin{figure}
    \centering
    \includegraphics[width=\textwidth]{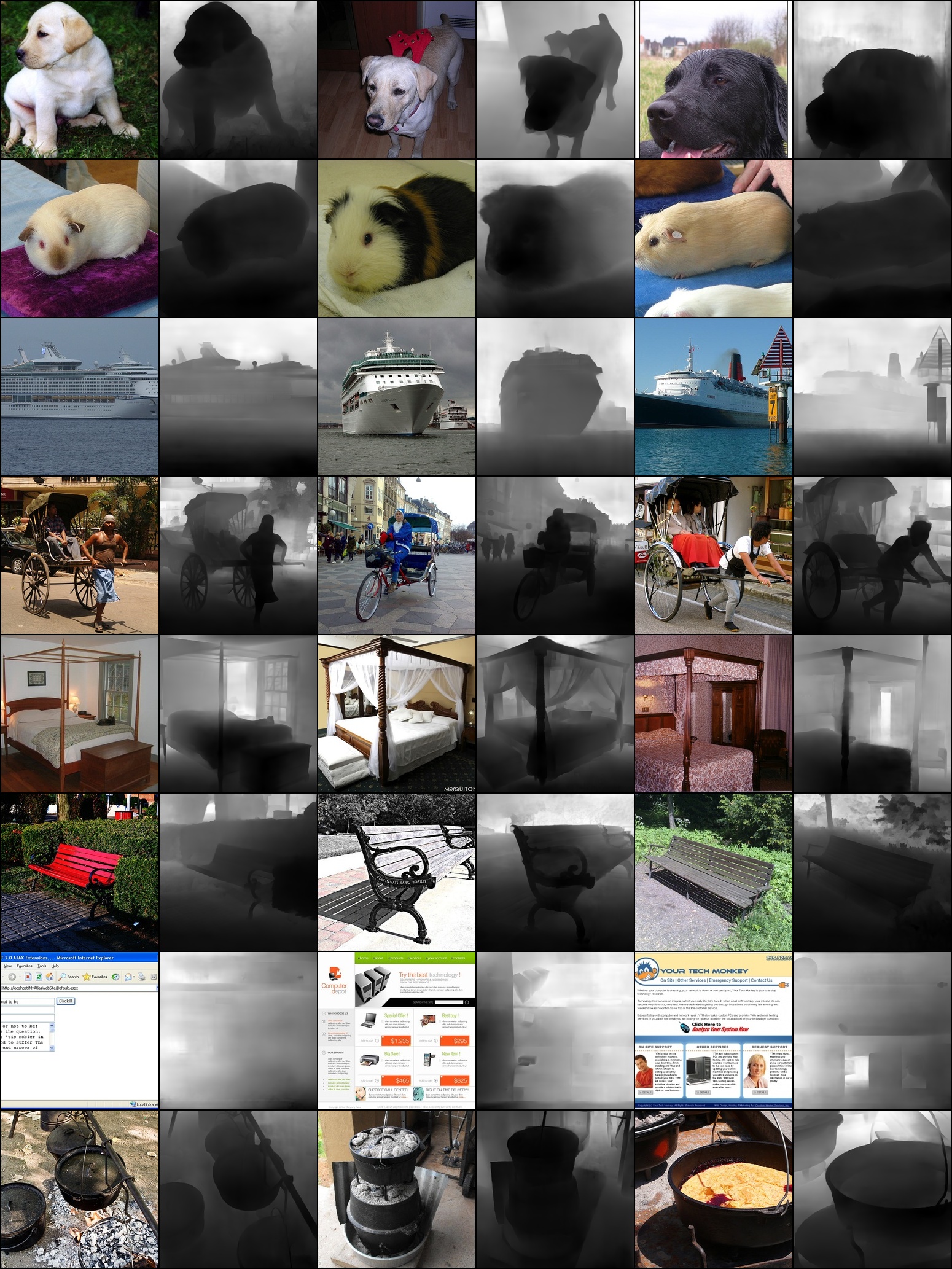}
    \caption{Depth maps on ImageNet $256^2$, predicted by LeReS~\citep{LeReS} depth estimator. See the generated depth by \modelname\ in \figref{fig:fake-depth}.}
    \label{fig:real-depth}
\end{figure}

\section{\metricfullname\ details}\label{ap:metric-details}

To detect and quantify the flatness of 3D generators, we propose a \metricfullname~(\metricname). To compute~\metricname, we sample $N$ latent codes with their corresponding radiance fields. For each radiance field, we perform integration following Eq.~\ref{eq:depth_rendering} to obtain its depth, however, we first set the $50\%$ of lowest density to zero. This is necessary to cull spurious density artifacts. 

We then normalize each depth map according to the corresponding near and far planes and compute a histogram with $B$ bins, showing the distribution of the depth values. To analyze how much the depth values are concentrated versus spread across the volume,  we compute its entropy for each distribution. Averaging over $N$ depth maps gives the sought score:
\begin{equation}
\small
\text{NFS} = \frac{1}{N}\sum_{i=1}^N\exp\left[ -\frac{1}{h \cdot w} \sum_{j=1}^B b(\bm d^{(i)})_j \right],
\end{equation}
where $b(\bm d^{(i)})_j$ is the normalized number of depth values in the $j$-th bin of the $i$-th depth map $\bm d^{(i)}$, with $N=256$  and $B=64$. \metricname~does not directly evaluate the quality of the geometry. Instead, it helps to detect and quantify how flat the generated geometry is. In \figref{fig:nfs}, we show examples of repetitive geometry generated by EG3D and a more diverse generation produced by 3DGP along with their depth histograms and {\metricname}.

Intuitively, NFS should provide high values to EG3D with a wide camera distribution, since its repetitive shapes are supposed to span the whole ray uniformly.
But surprisingly, it does not do so and is able to relfect this repeptitiveness artifact as well, yielding low scores, as can be observed from Tables~\ref{tab:2d-experiments} and \ref{tab:main-results}.

\section{Investigating the potential data leak from Depth Estimator}\label{ap:depth-prior}

LeReS~\citep{LeReS} depth estimator (which was used in our work) was pre-trained on a different set of images compared to ImageNet (or other animals datasets, explored in our work), and this theoretically could help our generator achieve better results by implicitly giving it access to a broader image set rather than giving only texture-less geometric guidance.
In this section, we investigate this, and conclude that it is not the case: the pre-training data of LeReS is too different from the explored datasets, containing almost no animal images --- while all the explored datasets are very animal-dominated, including ImageNet which contains 482 animal classes.
Hence, LeReS has no chance in helping our generator by giving it access to additional data.

\begin{figure}
    \centering
\includegraphics[width=\textwidth]{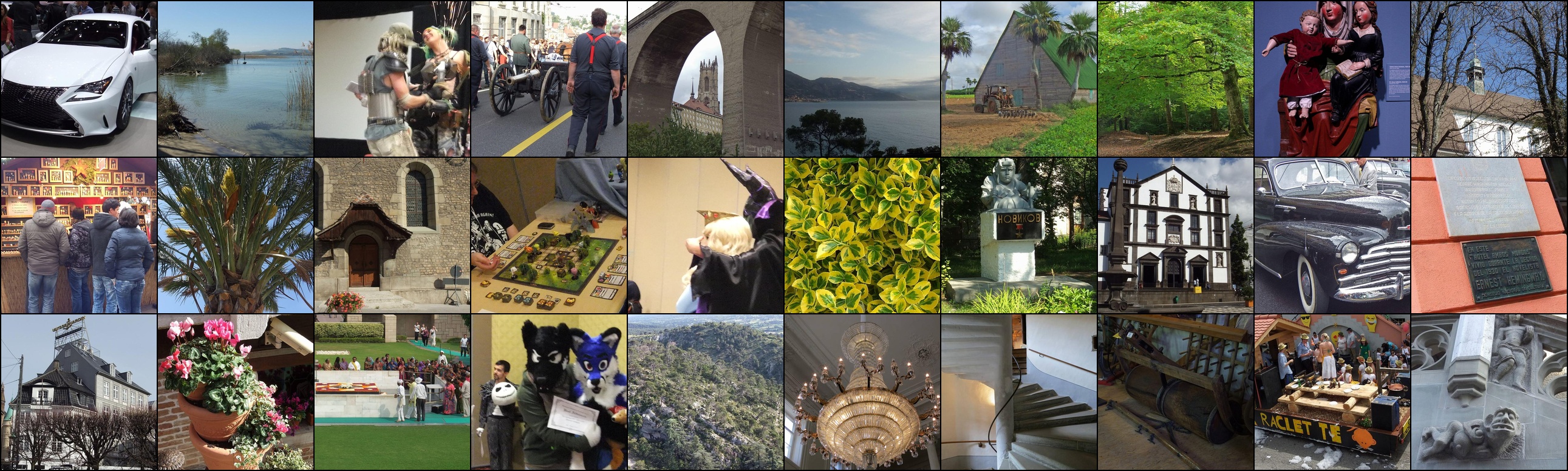}
\vfill
\includegraphics[width=\textwidth]{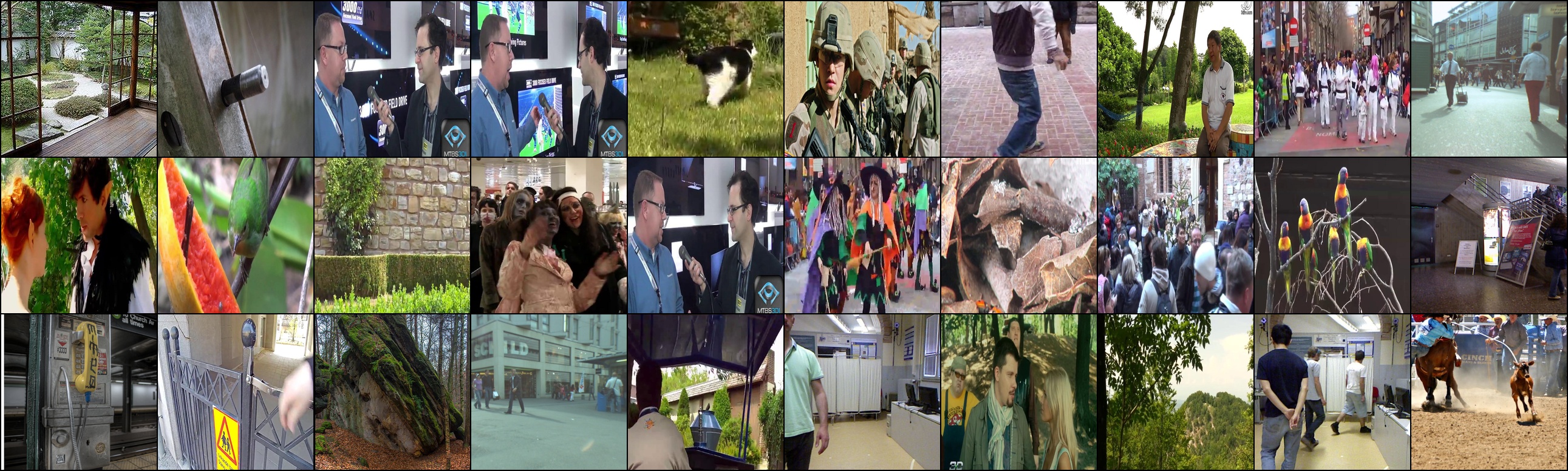}
\vfill
\includegraphics[width=\textwidth]{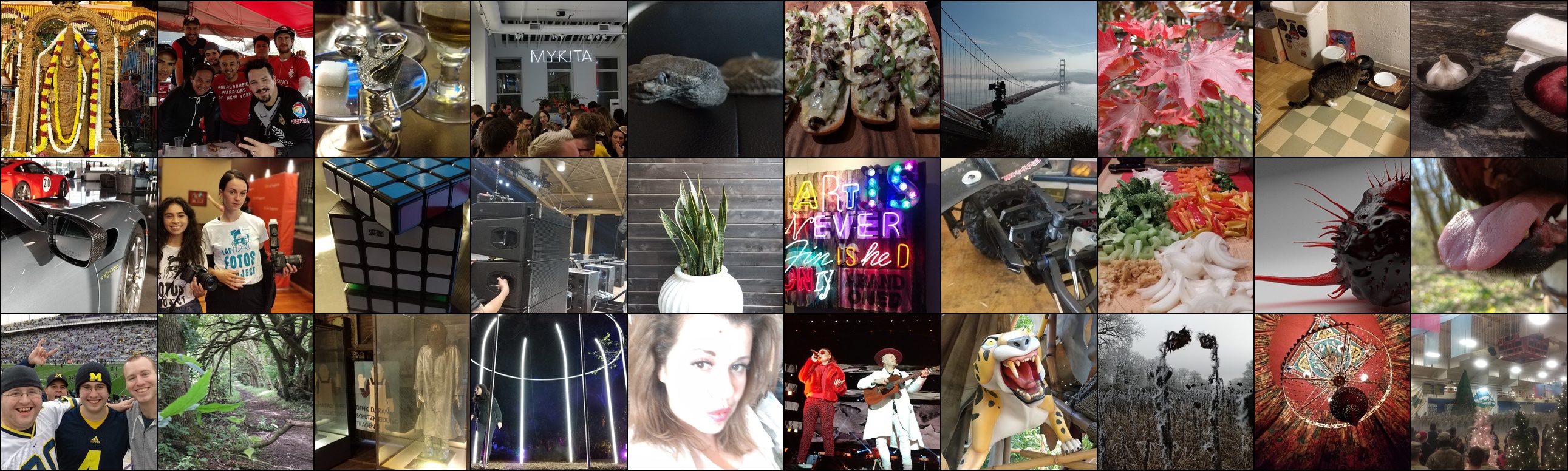}
\vfill
\includegraphics[width=\textwidth]{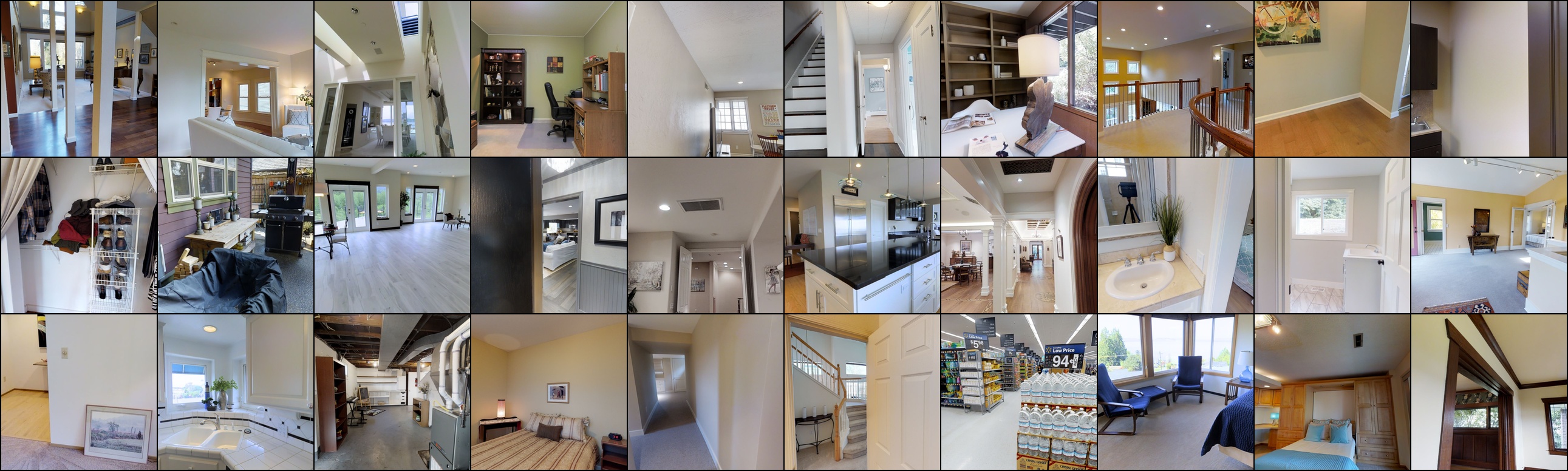}
    \caption{Random images from each pre-training dataset of the LeReS \citep{LeReS} depth estimator, used in our work. From top to bottom: HRWSI \citep{HRWSI}, DiverseDepth \citep{DiverseDepth}, Holopix50k \citep{Holopix50k}, and Taskonomy \citep{Taskonomy}. }
    \label{fig:leres-data}
\end{figure}

LeReS~\citep{LeReS} trains on a combination of 4 datasets (in fact, parts of them): 1) HRWSI~\citep{HRWSI} contains outdoor city imagery (e.g., buildings, monuments, landscapes); 2) DiverseDepth~\citep{DiverseDepth} contains clips from in-the-wild movies and videos; 3) Holopix50k~\citep{Holopix50k} — diverse, in-the-wild web images (it is the most similar to ImageNet in terms of underlying data distribution); and 4) Taskonomy~\citep{Taskonomy} contains indoor scenes (bedrooms, stores, etc.).
In \figref{fig:leres-data}, we provide 30 random images from each dataset.
In \tabref{tab:imagenet-animals}, we rigorously explore how many animal images does the pre-training data of LeReS contains.
We perform this in two ways: by directly counting the amount of animals with the pre-trained Mask R-CNN model~\cite{MaskRCNN} and by computing FID scores between the pre-training datasets of LeReS and ImageNet animal subset.
The pre-trained Mask R-CNN model from TorchVision is able to detect 10 animal classes: birds, cats, dogs, horses, sheeps, cows, elephants, bears, zebras and giraffes.
Following the official tutorial, we used a threshold of 0.8.
We downloaded the datasets with the official download scripts of LeReS.

\begin{table}[h!]
\caption{Investigating the distribution overlap between LeReS~\cite{LeReS} training data and the animals subset of ImageNet.}
\label{tab:leres-data}
\centering
\resizebox{1.0\linewidth}{!}{
\begin{tabular}{lccccc}
\toprule
\multirow{2}{*}{Dataset} & \multirow{2}{*}{\#images} & \multicolumn{2}{c}{Animal subset} & \multicolumn{2}{c}{FID} \\
& & \#images & \%images & ImageNet & ImageNet$_\text{animals}$ \\
\midrule
HRWSI~\cite{HRWSI} & 18.2k & 1.05k & 5.75\% & 71.48 & 97.79 \\
DiverseDepth~\cite{DiverseDepth} & 95.4k & 7.9k & 8.26\% & 90.99 & 122.6 \\
Holopix50k~\cite{Holopix50k} & 42k & 4k & 9.56 & 47.44 & 81.62 \\
Taskonomy~\cite{Taskonomy} & 134.7k & 0.5k & 0.36\% & 135.1 & 154.3 \\
\midrule
All pre-training data & 290k & 9.4k & 3.79\% & 77.46 & 108.6 \\
\midrule
ImageNet & 1281.2k & 618.7k & 48.3\% & 0.0 & 19.06 \\
\bottomrule
\end{tabular}
}
\end{table}

From \tabref{tab:leres-data}, one can see that the pre-training data of LeReS contains just 3.79\% of animal images, while ImageNet contains 48.3\%.
In this way, it is too far away in terms of distribution from the Animal subset of ImageNet.
From this, one can conclude the following: \emph{Depth Estimator has almost never seen animals during training, since its training datasets have 600 fewer animal images than ImageNet.}
This means that our generator does not receive any unfair advantage in synthesizing animals from using adversarial depth supervision by implicitly getting access to a larger set of images.

How does it affect our generator's performance in synthesizing animals?
Will it have an unusually poor FID compared to non-animal data or compared to other methods?
In \tabref{tab:imagenet-animals}, we report FID scores of each generator on animal vs non-animal subsets of ImageNet.
From these scores, one can see that the trend is the same as for FID on full ImageNet.
This implies that our generator performs equally well on data which constitutes the main part of our training dataset and is not a part of the depth estimator pre-training.

\begin{table}
\caption{FID scores of different generators on 482 animal classes of ImageNet $256^2$ (each generator was trained on all the classes of ImageNet $256^2$). For this evaluation, we used the images only from the animal classes for both real and fake data.}
\label{tab:imagenet-animals}
\centering
\begin{tabular}{lcccc}
\toprule
Method & Synthesis type & Animal FID $\downarrow$ & Non-Animal FID $\downarrow$ & FID $\downarrow$ \\
\midrule
StyleGAN-XL & 2D & 4.53 & 4.81 & 2.30 \\
\midrule
EG3D (wide camera) & 3D-aware & 44.55 & 49.14 & 25.6 \\
EpiGRAF & 3D & 78.65 & 81.38 & 47.56 \\
\modelname~(ours) & 3D & 47.32 & 50.94 & 26.47 \\
\bottomrule
\end{tabular}
\end{table}

In this way, one can conclude that adversarial depth supervision helps to achieve better synthesis quality only through geometric guidance.
It does not help the generator to do this by leaking the knowledge of pre-trained data from the pre-trained depth estimator.

\section{Geometry evaluation on ShapeNet}\label{ap:geometry-evaluation}

In this section, we present the details and the results of an additional study, which rigorously shows that our proposed adversarial depth supervision improves the geometry quality.

\subsection{Rendering details}

\begin{figure}
    \centering
    \includegraphics[width=\textwidth]{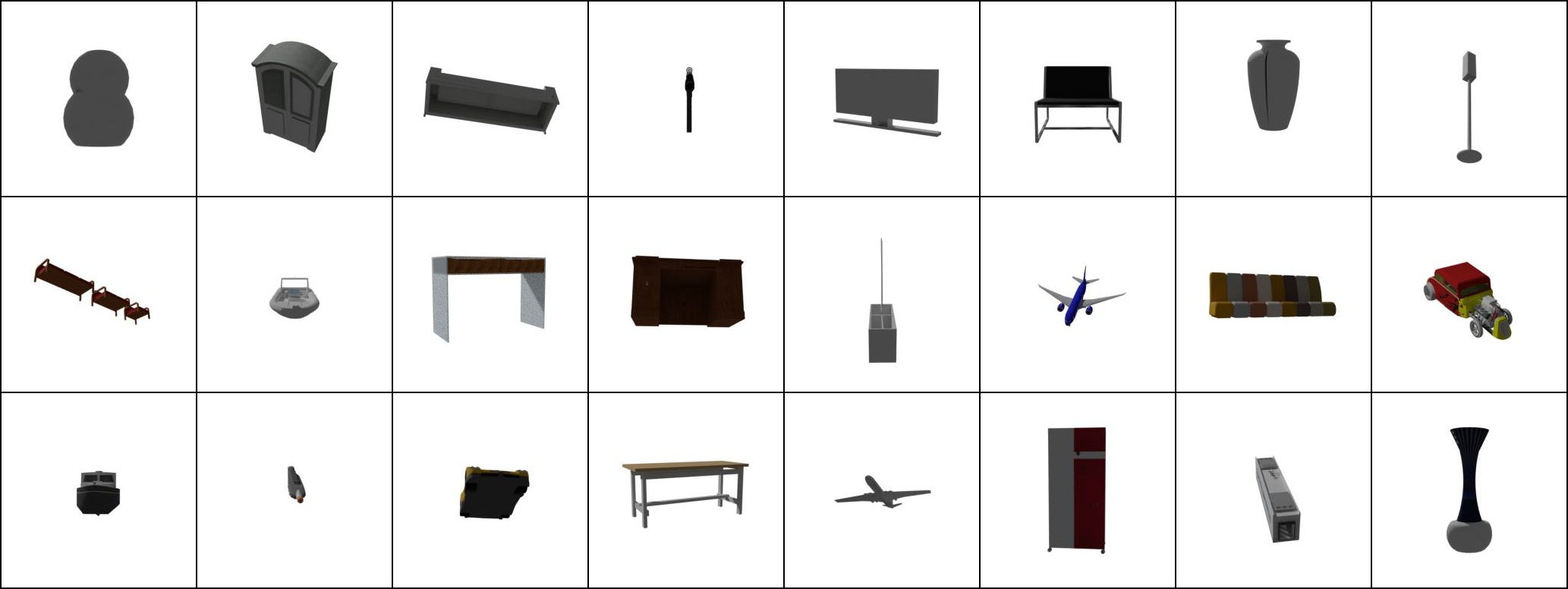}
    \caption{Examples of real images for our rendered ShapeNet $256^2$ dataset.}
    \label{fig:shapenet-real}
\end{figure}

We take the ShapeNet dataset~\cite{ShapeNet}, which consists of ~50k models from 54 classes and render it from random \emph{frontal} camera positions.
For this, we set the camera on a sphere of radius $2$ and randomly choose its position by sampling rotation and elevation angles from $N(0, \pi/8)$ and $N(\pi/2, \pi/8)$, respectively.
We chose frontal views to be closer to the real-world scenario: the most popular image synthesis datasets are dominated by the frontal views.
Also, if the dataset is not frontal, then there is less necessity in additional geometry supervision: it is a good enough 3D bias (which does not exist in modern in-the-wild datasets).
For wide camera distribution, a generator can learn proper geometry on its own: see experiments on synthetic datasets with 360 degrees camera coverage by \cite{EpiGRAF}.
We render just a single view per model since this is also the scenario which we typically have in real datasets (e.g., in all our explored datasets, there is just a single view per scene).
Some of the models in ShapeNet are broken (they lack the corresponding mesh files), so we discard them.
In total, this gave us 51209 training images, which are visualized in \figref{fig:shapenet-real}.
We used BlenderProc~\cite{BlenderProc} to do rendering.
Also, during rendering we removed the transparent elements of the meshes since they were providing aliasing issues in the depth maps.

\subsection{Experiments}

\textbf{Experimental setup}.
After rendering, we trained our main baselines, EG3D~\citep{EG3D} and EpiGRAF~\citep{EpiGRAF}, on this dataset using the ground truth camera parametrization and distribution when sampling the camera poses.
For \modelname, we trained it in the following variants: 1) with the default adversarial depth supervision (i.e., using $\depthprob = 0.5$); and 2-3) with the adversarial depth supervision, but with corrupted depth maps, where corruption is simulated by blurring, similar to \citep{GSN}.
Also, we trained StyleGAN2~\citep{StyleGAN2} as a lower bound on FID.
We disabled camera distribution learning and knowledge distillation for the experiments with \modelname\ to avoid unnecessary complications: we investigate only adversarial depth supervision in this study.
For all the baselines, we use white background in volumetric rendering.

\textbf{Evaluation}.
We compare the methods in terms of FID~\citep{FID} and also Frechet Pointcloud Distance (FPD), proposed by \cite{FPD}.
This metric is an FID analog for point clouds: it uses a pre-trained PointNet~\citep{PointNet} to extract the features from point clouds and then computes Frechet distance between real and fake representations sets, approximating their underlying distribution as multi-variate normal one.

We used the official codebase of \cite{FPD} to compute the metrics.
In their original work, \cite{FPD} used a customly trained PointNet~\citep{PointNet} to extract the features.
But we observed an issue with it: it was sometimes producing unusually high FPD scores in the order of $10^6$, even for real point clouds.
This is why we extracted pointclouds features with a pre-trained PointNet++~\citep{PointNet++} model from a popular public implementation~\citep{Pytorch_Pointnet_Pointnet2}.

To extract point clouds from the models, we first extracted the surfaces from them via marching cubes.
For this, we sampled density fields in $256^3$ resolution, and then used the marching cubes implementation of PyMCubes~\citep{PyMCubes} to extract the surfaces.
Following EG3D, we thresholded the surface for marching cubes at the density value of $\sigma=10$.
Our overall pipeline is identical to the original procedure of SDF extraction in the EG3D repo, but simpler in terms of implementation.
After that, we extracted point clouds by sampling 2,048 points on the surface for both real and fake meshes using uniform sampling from trimesh~\cite{Trimesh}.

\subsection{Results}

\begin{figure}
\centering
    \includegraphics[width=\textwidth]{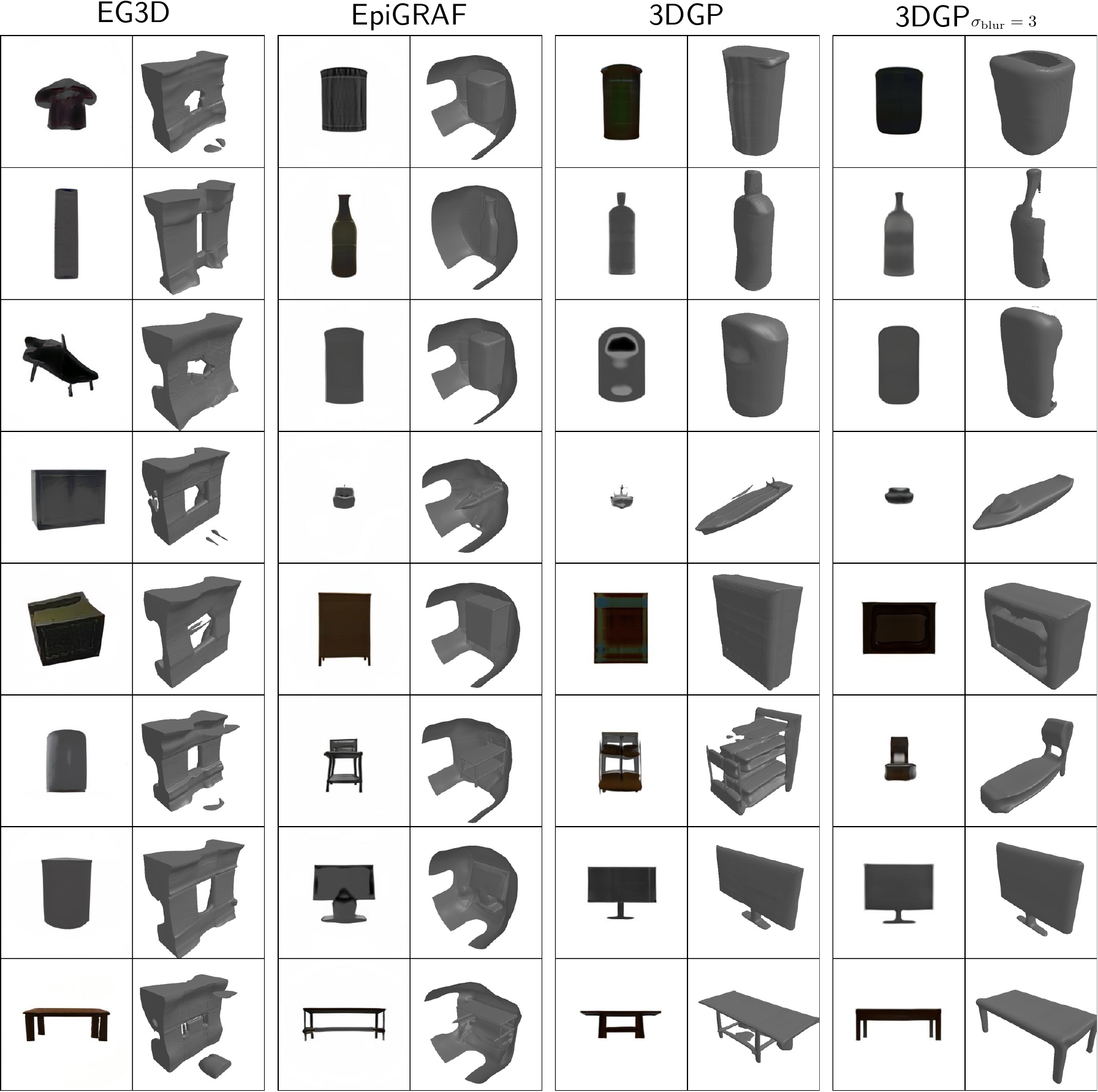}
    \caption{Random samples (seed 1, classes 1-8) for ShapeNet $256^2$ for our method and the baselines.}
    \label{fig:shapenet-samples}
\end{figure}

The results of these experiments are presented in \tabref{tab:shapenet-results}.
We also provide random sample examples (seed 1, classes 1-8) in \figref{fig:shapenet-samples}.
EG3D~\citep{EG3D} fails to recover the geometry and generates inverted shapes with hollow geometry.
This is highlighted by its very high FPD score.
EpiGRAF~\citep{EpiGRAF} is able to recover shapes, but it models the white background via an additional sphere, which also damages its FPD.
Our method, in contrast, learns proper geometry and correctly drops the background.
When the depth maps are corrupted with blurring, it deteriorates both its geometry quality and texture quality, which is also confirmed by the metrics and the provided samples.

\begin{table}
\caption{Geometry evaluation for different models on ShapeNet $256^2$~\citep{ShapeNet}.}
\label{tab:shapenet-results}
\centering
\begin{tabular}{lcccc}
\toprule
Method & Synthesis type & FID $\downarrow$ & FPD $\downarrow$ \\
\midrule
EG3D & 3D-aware & 17.22 & 2770.5 \\
EpiGRAF & 3D & 21.58 & 424.0 \\
\modelname~\ours & 3D & 14.38 & 80.67 \\
~with $\sigma_\text{blur} = 1$ & 3D & 18.33 & 106.9 \\
~with $\sigma_\text{blur} = 3$ & 3D & 30.37 & 139.5 \\
~with $\sigma_\text{blur} = 10$ & 3D & 99.41 & 807.4 \\
\midrule
StyleGAN2 & 2D & 5.54 & N/A & \\
\bottomrule
\end{tabular}
\end{table}

\section{Details on the experiments with 3D Photo Inpainting}\label{ap:3d-photo}

In this section, we provide the details of our experiments on combining 2D generation with 3D Photo Inpainting techniques.

For a 2D generator, we chose StyleGAN-XL~\citep{StyleGAN-XL} since it achieves state-of-the-art visual quality on ImageNet, as measured by FID~\citep{FID}.
We took the original checkpoint for $256^2$ generation from the official repository.\footnote{\url{https://github.com/autonomousvision/stylegan_xl}}.
First, we generated 50,000 images with the model and computed their FID: this gave the value of 2.51 vs 2.26, reported by the authors, which is a negligible difference.

After that, we used the original codebase of 3DPhoto~\citep{3DPhoto} to synthesize the 3D variations of 10k random images.
We kept all the hyperparameters the same.
In 3DPhoto, there is no spherical camera parametrization, this is why we had to simulate it the following way.
We assumed that the sphere center lies at the median depth value from the original camera position (which is $(0, 0, 0)$), and had been rotating the camera around that point.
As discussed in \secref{sec:experiments}, we used the narrow camera distribution to sample the points.
Namely, we used the normal distributions with standard deviations of $\sigma_\text{yaw}=0.3$ and $\sigma_\text{pitch}=0.15$ for sampling rotation and elevation angles, respectively.

\begin{figure}
\centering
\includegraphics[width=\textwidth]{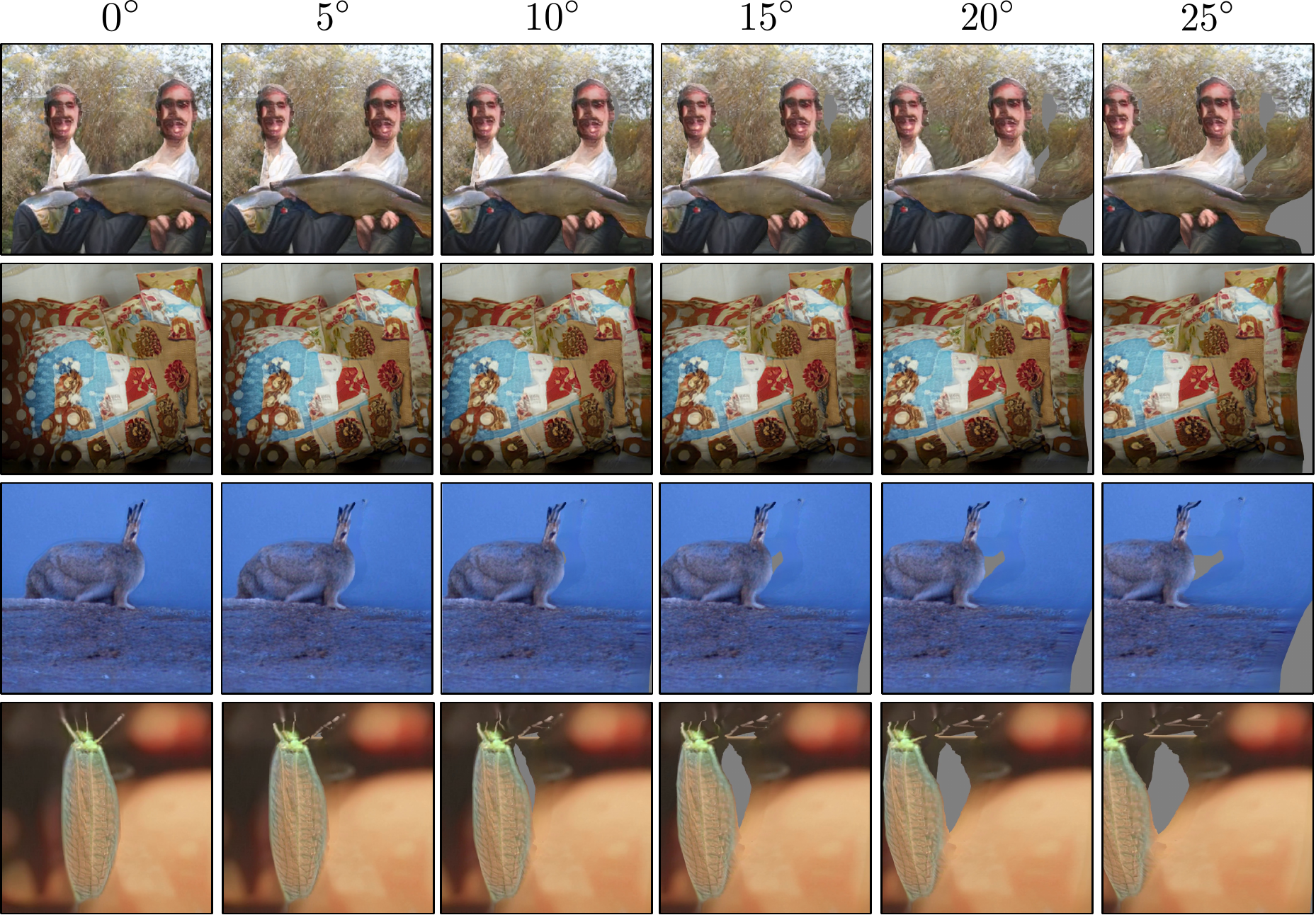}
\caption{Random samples from StyleGAN-XL~\citep{StyleGAN-XL} paired with 3D Photo Inpainting~\citep{3DPhoto}. For large camera movements, there appear severe artifacts, like gray areas and interpolation artifacts.}
\label{fig:3d-photo}
\end{figure}

For negligible camera variations, this strategy produces excellent results.
While it is not perfectly view consistent, it inherits state-of-the-art image quality from the 2D generator, which was used to synthesize the original images.
But for larger camera variations, the quality becomes to deteriorate very quickly.
Not only noticeable inpainting artifacts start to appear, but also outpainting problem emerges.
The model was not trained to perform image extrapolation or inpaint large regions and fails to fill the holes which appear in larger camera movements.
We provide the visualizations for it in \figref{fig:3d-photo}.

\section{Entropy regularization}\label{ap:entropy-reg}

After the submission, we explored another strategy to regularize the camera generator and observed that it is more flexible and easier to tune.
Intuitively, the strategy is to maximize the entropy of each predicted camera parameter $\phi_i$:
\begin{equation}
\Lphi = H(\varphi_i) = \expect[p_G(\varphi_i)]{-\log p_G \varphi_i},
\end{equation}
where $p_G(\varphi_i)$ is the camera generator's distribution over $\varphi_i$, $H$ is differential entropy.

To do this, we used the POT package~\cite{POT} to minimize the Earth Mover's Distance with the uniform distribution:
\begin{equation}
\Lphi = \text{EMD}(p_G(\varphi_i), U[m_{\varphi_i}, M_{\varphi_i}]),
\end{equation}
where $\text{EMD}(P, Q)$ denotes the Earth Mover's Distance between the distributions $P$ and $Q$, $U[a, b]$ is the uniform distribution on $[a, b]$ and $m_{\varphi_i}, M_{\varphi_i}$ are minimum and maximum values for a given camera parameter $\varphi_i$.

In practice, we used 64 samples to approximate the EMD and a non-regularized Wasserstein distance.

\end{document}